\documentclass[10pt,twocolumn,letterpaper]{article}

\usepackage[pagenumbers]{cvpr} 

\usepackage[accsupp]{axessibility}
\usepackage{graphicx}
\usepackage{amsmath}
\usepackage{amssymb}
\usepackage{wasysym}
\usepackage{booktabs}
\usepackage{multirow}
\usepackage{colortbl}
\usepackage{bm}

\usepackage[noend]{algpseudocode}
\usepackage{algorithmicx,algorithm}

\usepackage[pagebackref,breaklinks,colorlinks]{hyperref}

\usepackage[capitalize]{cleveref}
\crefname{section}{Sec.}{Secs.}
\Crefname{section}{Section}{Sections}
\Crefname{table}{Table}{Tables}
\crefname{table}{Tab.}{Tabs.}

\def\cvprPaperID{2301} 
\def\confName{CVPR}
\def\confYear{2022}

\begin{document}


\title{Automatic Relation-aware Graph Network Proliferation}

\author{
    Shaofei Cai$^{1,2}$, Liang Li$^{1}$\thanks{Corresponding author.}, Xinzhe Han$^{1,2}$, Jiebo Luo$^{3}$, Zheng-Jun Zha$^{4}$, Qingming Huang$^{1,2,5}$ \\
    $^{1}$Key Lab of Intell. Info. Process., Inst. of Comput. Tech., CAS, Beijing, China \\
    $^{2}$University of Chinese Academy of Sciences, Beijing, China, $^{3}$University of Rochester \\
    $^{4}$University of Science and Technology of China, China, $^{5}$Peng Cheng Laboratory, Shenzhen, China \\
    {\tt \small \{shaofei.cai,xinzhe.han\}@vipl.ict.ac.cn,liang.li@ict.ac.cn,} \\
    {\tt \small jluo@cs.rochester.edu,zhazj@ustc.edu.cn,qmhuang@ucas.ac.cn}
}

\maketitle

\begin{abstract}
Graph neural architecture search has sparked much attention as Graph Neural Networks (GNNs) have shown powerful reasoning capability in many relational tasks. 
However, the currently used graph search space overemphasizes learning node features and neglects mining hierarchical relational information. 
Moreover, due to diverse mechanisms in the message passing, the graph search space is much larger than that of CNNs. 
This hinders the straightforward application of classical search strategies for exploring complicated graph search space. 
We propose Automatic Relation-aware Graph Network Proliferation (ARGNP) for efficiently searching GNNs with a relation-guided message passing mechanism. 
Specifically, we first devise a novel dual relation-aware graph search space that comprises both node and relation learning operations. 
These operations can extract hierarchical node/relational information and provide anisotropic guidance for message passing on a graph. 
Second, analogous to cell proliferation, we design a network proliferation search paradigm to progressively determine the GNN architectures by iteratively performing network division and differentiation. 
The experiments on six datasets for four graph learning tasks demonstrate that GNNs produced by our method are superior to the current state-of-the-art hand-crafted and search-based GNNs. 
Codes are available at \url{https://github.com/phython96/ARGNP}. 

\end{abstract}


\section{Introduction}
Graph neural networks (GNNs), as a dominant paradigm to handle graph-structured data, have significantly promoted the performance in many relation reasoning tasks, such as molecular prediction \cite{Bouritsas2020ImprovingGN, Hussain2021EdgeaugmentedGT, Gilmer2017NeuralMP, Dwivedi2020BenchmarkingGN}, social network analysis \cite{Liu2021ContentMA, Li2021RelevanceAwareAU}, 3D point cloud recognition \cite{landrieu2018large, qi20173d, DGCNN, SGAS}, object detection \cite{hu2018relation, gu2018learning}, semantic segmentation \cite{Li2020SpatialPB, Hu2020ClasswiseDG, Wang2019GraphAC}, few-shot learning \cite{Yang2020DPGNDP, Satorras2018FewShotLW, Kim2019EdgeLabelingGN}, \etc. 
Despite their great success, the architectures of GNNs are usually manually designed, which requires tremendous expert knowledge and intensive trial and error. 
To explore advanced GNN architectures and reduce the human intervention, researchers attempt to automate the design process with the help of neural architecture search (NAS) \cite{Zhou2019AutoGNNNA, Li2021OneshotGN, Gao2020GraphNA, Pan2021AutoSTGNA, Zhao2020SimplifyingAS} and have achieved superior performance. 
This is known as graph neural architecture search, where there are two most critical components: (1) graph search space and (2) search strategy. 


The graph search space defines which graph neural networks can be represented in principle, determining the upper bound of networks' reasoning capability. 
Current graph search space mainly focuses on designing node-learning operations, which is categorized into macro search space \cite{SGASSG, Pan2021AutoSTGNA} and micro search space \cite{Zhou2019AutoGNNNA, Cai2021RethinkingGN, Li2021OneshotGN, Gao2020GraphNA}. 
The former explores the combinations of existing message passing mechanisms (\ie, using general-purpose GNNs as candidate operations), while the latter emphasizes the construction for novel ones (\ie, designing fine-grained operations such as node aggregating and feature combining functions). 
However, they all neglect mining the latent hierarchical relational information associated with edges. 
In fact, relational information can provide anisotropic guidance for message aggregation of neighboring nodes, which is critical for constructing relation-guided message passing mechanisms. 
Motivated by this, in this paper, we explore designing micro graph search space from the perspective of learning both hierarchical relation and node features. 

The search strategy details how to explore search space, determining the search efficiency and effect. 
Some early works \cite{Gao2020GraphNA, Zhou2019AutoGNNNA, Zhao2020SimplifyingAS } apply reinforcement learning (RL) based strategy to search for GNNs by building, training, and evaluating various graph neural architectures from scratch, which is extremely time-consuming. 
Recently, due to the high computational efficiency, one-shot differentiable strategies \cite{SGASSG, Li2021OneshotGN, Cai2021RethinkingGN} have attracted a lot of interest, which consists of three stages: supernet training, subnet searching, and subnet retraining. 
They boost the search efficiency from the parameter sharing among subnets and supertnet. 
Using the auxiliary supernet can avoid training each child graph neural architecture individually, but this may cause severe subnet interference \cite{Zhang2020DeeperII, Zhang2021AceNASLT}. 
Besides, it is limited in searching large GNN architectures due to the quadratic complexity of storing and training the supernet. 
As a compromise, researchers introduce the cell trick where the architecture is a stack of several same building blocks \cite{SGASSG, Li2021OneshotGN, Cai2021RethinkingGN, DARTS}. 
This shifts the searching objective from the whole architecture to small cells but seriously narrows the original search space. 
The above limitations, \ie the subnet interference, the high space-time complexity and the shrink of search space, bring a severe negative impact for graph neural architecture search. 



\begin{figure*}[t]
    \centering
    \includegraphics[scale = 0.92, trim = 72 580 20 64, clip]{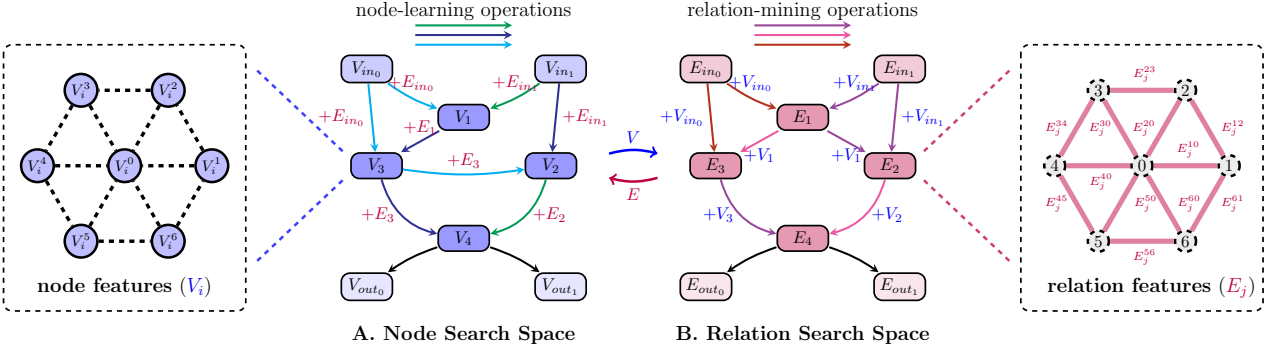}
    \vspace{-1em}
    \caption{
        \textbf{Our dual relation-aware graph search space. }
        It comprises node and relation search space. 
        The vertex $\bm{V}_{i}$ and vertex $\bm{E}_{j}$ denote a group of node and relation features on the graph, respectively. 
        \textcolor{red}{$+\bm{E}_{j}$} means that learning node features require the guidance of relation features. 
        \textcolor{blue}{$+\bm{V}_{i}$} means that the node features are required when extracting relational information. 
    }
    \vspace{-0.8em}
    \label{space_figure}
\end{figure*}

In this paper, we propose the Automatic Relation-aware Graph Network Proliferation (ARGNP) to efficiently search the optimal GNN architectures with a relation-guided message passing mechanism. 
First, we design a dual relation-aware graph search space comprising both relation and node search space, as shown in Figure \ref{space_figure}. 
The relation search space introduces diverse relation-mining operations to extract relational information hidden in edge-connected nodes. 
It allows arbitrary valid connection modes among relation-mining operations and forms the hierarchical relation-learning structure. 
Different connection modes result in the group of relation features with different message-passing preferences, which favors different graph tasks. 
The node search space defines a series of node-learning operations which implements the anisotropic message aggregation under the guidance of relation features. 

\begin{figure}[t]
    \centering
    \includegraphics[scale = 0.620, trim = 110 355 0 330, clip]{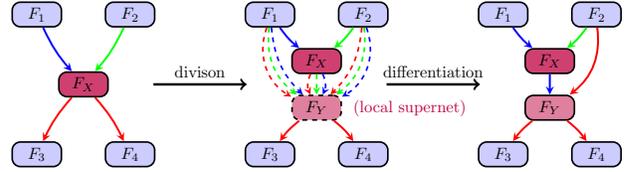}
    \vspace{-1.0 em}
    \caption{
        \textbf{One iteration of the proposed network proliferation search paradigm.} 
        $F_{(\cdot)}$ denotes the intermediate feature vertex in an architecture. 
        The edges with different colors are associated with different operations. 
        $F_{Y}$ is the newly divided part of $F_{X}$, which joins $F_{1}$, $F_{2}$ and $F_{X}$ to build a local supernet. 
    }
    \vspace{-1.0 em}
    \label{fission_intro}
\end{figure}

Second, analogous to cell proliferation, we devise a novel search paradigm called \emph{network proliferation} to progressively explore the graph search space. 
Instead of directly optimizing the global supernet, we search the final graph neural architecture by iteratively performing network division and network differentiation. 
Figure \ref{fission_intro} shows one iteration process. 
During network division, each intermediate feature vertex is divided into two parts. 
One retains original operations and connections while the other builds a local supernet. 
Network differentiation aims to differentiate the local supernet into a specific subnet. 
Theoretically, we proved that such a search paradigm achieves the linear space-time complexity. 
This enables our search to thoroughly free from the cell trick. 
The network proliferation decomposes the training of global supernet into sequential local supernets optimization, which alleviates the interference among child graph neural architectures.

Our contributions are summarized as follows: 
\textbf{(1)} A novel dual relation-aware graph search space comprising both node and relation search space. 
It can derive GNNs with a relation-guided message passing mechanism. 
\textbf{(2)} A network proliferation search paradigm. It sequentially performs network division and differentiation to explore the optimal GNN architecture within linear space-time complexity. 
\textbf{(3)} Experiment results on six datasets for four classical graph learning tasks show that our method outperforms human-crafted and other search-based GNNs by a large margin. The code will be released publicly. 

\section{Related Work}

\textbf{Graph Neural Networks (GNNs)} have been successfully applied to operate on the graph-structure data~\cite{GG, GNNFILM, PNA, GCN, GAT, GIN, FGCN, SAGE}. 
Current GNNs are constructed on message passing mechanisms and can be categorized into two groups, isotropic and anisotropic. 
Isotropic GNNs~\cite{SAGE, GCN, GIN} aim at extending the original convolution operation to graphs. 
Anisotropic GNNs enhance the original models with anisotropic operations on graphs~\cite{Edge}, such as gating and attention mechanism~\cite{GG, AA1, AA2, GAT, FGCN}. 
Anisotropic methods usually achieve better performance, since they can weigh the importance of edges according to the node features and implement the adaptive feature aggregation. 

\textbf{Neural Architecture Search (NAS)} aims at automatically finding the optimal neural architectures specific to dataset~\cite{NASRL, Proxy, DARTS, DARTS-, SGAS, DARTS+, DARTS, NAS_SHARE, SNAS, NAS16}. 
Search space and strategy are the most essential components in NAS. 
Search space defines which architectures can be represented in principle. 
Search strategy details how to explore the search space. 
Methods can be mainly categorized into three groups, \ie, reinforcement learning (RL)~\cite{NASRL, NAS16, NAS_SHARE}, evolutionary algorithms (EA)~\cite{EA1, EA2, EA3} and gradient-based (GB)~\cite{DARTS, Xu2019PCDARTSPC, DARTS-, DARTS+, SGAS, Proxy}. 
Benefiting from the high efficiency, gradient-based differentiable search strategies attracted the attention of increasing researchers. 

\textbf{Graph Neural Architecture Search (GNAS)} is proposed to automatically find the best GNNs for the given specific graph task. 
The current GNAS methods~\cite{Cai2021RethinkingGN, Zhou2019AutoGNNNA, Li2021OneshotGN, Gao2020GraphNA, Pan2021AutoSTGNA, Zhao2020SimplifyingAS} mainly focus on designing graph search space by introducing neighbor aggregation, activation functions, \etc, but neglect the importance of relation mining. 
To our best knowledge, we are the first to take mining relational information into account during devising graph search space. 


\section{Method}
\subsection{Preliminaries}

A graph neural architecture uses graph-structured data as input and outputs high-dimensional node and edge features (relation features). 
The input data can be represented as $\{\bm{G}, \bm{V}_{in}, \bm{E}_{in}\}$, where $\bm{G}$ is the input graph structure with $n$ nodes and $m$ edges, $\bm{V}_{in} \in \mathbb{R}^{n \times d_{V}}$ denotes the input node features, $\bm{E}_{in} \in \mathbb{R}^{m \times d_{E}}$ denotes the input edge features, $d_V$ and $d_E$ denote the feature dimension. 
Notably, if there is no original edge feature in the dataset, we initalize $\bm{E}_{in}$ with $[1, \cdots, 1] \in \mathbb{R}^{m \times 1}$ for subsequent relation learning. 
Our dual relation-aware graph search space comprises both node and edge search space. 
The derived computation structure of node or relation search space is a directed acyclic graph (DAG). 
To avoid confusion, the node and edge in the DAG is renamed to ``vertex'' and ``link'', where ``vertex'' denotes an intermediate features ($\bm{V}_{i}$ or $\bm{E}_{j}$), the ``link'' is associated with an operation $\bm{o}$ or a mixture operation $\bm{\bar{o}}$.

\subsection{Relation-aware Graph Search Space}


In this subsection, we detail how our dual relation-aware graph search space is devised. 
As shown in Figure \ref{space_figure}, it comprises node search space and relation search space. 
These two spaces are not individual in essence, instead, they communicate information with each other. 
The relational information is extracted from node features. 
In turn, it provides anisotropic guidance for learning better node features. 

\noindent 
\textbf{Node search space.}
The computation structure derived by node search space is a directed acyclic graph, which consists of an ordered sequence of $N + 4$ vertices, which constitutes the set 
$\{\bm{V}_{in_0}, \bm{V}_{in_1}, \bm{V}_{1}, \cdots, \bm{V}_{N}, \bm{V}_{out_0}, \bm{V}_{out_1}\}$. 
Each vertex is a latent representation denoted as $\bm{V}_{i}$ (\ie, node features in a GNN layer, as shown in the left of Figure \ref{space_figure}), where $i$ is its topological order in the DAG. 
In order to be compatible with the cell trick, we introduce two input vertices and two output vertices. 
$\bm{V}_{in_0}$ and $\bm{V}_{in_1}$ are the outputs of previous two cells while $\bm{V}_{out_0}$ and $\bm{V}_{out_1}$ are the inputs of post two cells. 
In fact, our method can thoroughly free from the cell trick. 
In this situation, we have $\bm{V}_{in_0} = \bm{V}_{in_1}, \bm{V}_{out_0} = \bm{V}_{out_1}$. 
Each directed link $(i, j)$ in the DAG is associated with one node-learning operation $\bm{o}_{\mathcal{V}}^{(i,j)}$, that transforms the information from $\bm{V}_{i}$ to $\bm{V}_{j}$, guided by relation features $\bm{E}_{i}$. 
The transformed information is denoted as $\bm{V}_{i \rightarrow j} = \bm{o}_{\mathcal{V}}^{(i, j)}(\bm{V}_{i}, \bm{E}_{i}, f_{i,j})$, 
where $f_{i,j}$ is an aggregating function which takes multiset as input, such as $mean$, $max$, $sum$, $std$, \etc. 
We leverage feature-wise linear modulation (FiLM) mechanism \cite{GNNFILM} to implement the anisotropic guidance of message propagation.
This allows the model to dynamically up-weight and down-weight node features based on the relational information. 
Specifically, we compute an element-wise affine transformation $\bm{\gamma}$ and $\bm{\beta}$ based on the input relation features $\bm{E}_{i}$ and use it to modulate the incoming messages during message passing. 
Given a node $t$ on the graph, its transformed feature $\bm{V}_{i \rightarrow j}^{t}$ is formulated as follows
\begin{gather}
    \begin{split}
    \bm{V}_{i \rightarrow j}^{t} = f_{i,j}(\{\bm{\gamma}_{s,t} \odot &\bm{V}_{i}^{s} + \bm{\beta}_{s,t} | s \in \mathcal{N}(t)\}), \\
    \bm{\gamma}_{s,t}, \bm{\beta}_{s,t} &= g(\bm{E}_{i}^{s,t}; \bm{\theta}), 
    \end{split}
\end{gather}
where $g(\cdot)$ is a two-layer multilayer perceptron to compute affine transformation with the learnable parameters $\bm{\theta}$, $\mathcal{N}(t)$ denotes the set of neighbors of target node $t$ on the graph. 
Following previous works \cite{SGASSG, Li2021OneshotGN, Cai2021RethinkingGN, DARTS}, we allow two inputs for each intermediate vertex, \ie, $\bm{V}_{i} = \bm{V}_{p_1 \rightarrow i} + \bm{V}_{p_2 \rightarrow i}$, where $p_1,p_2 \in \{in_0, in_1, 1, 2, \cdots, i-1\}$. 
For each node-learning operation, its aggregating function is optional. 
Different aggregating function $f_{i,j}$ captures different types of information. 
For example, $sum$ captures the structural information \cite{GIN}, $max$ captures the representative information, $mean$ and $std$ captures the statistical information from the neighboring nodes. 
The node search space has 8 candidate node-learning operations: \emph{V\_MAX}, \emph{V\_SUM}, \emph{V\_MEAN}, \emph{V\_STD}, \emph{V\_GEM2}, \emph{V\_GEM3}, \emph{skip-connect}, and \emph{zero} operation. 
We detail all eight node-learning operation options in the {supplementary material}. 

\begin{figure*}[t] 
    \centering
    \includegraphics[scale=0.9, trim = 50 325 0 300, clip]{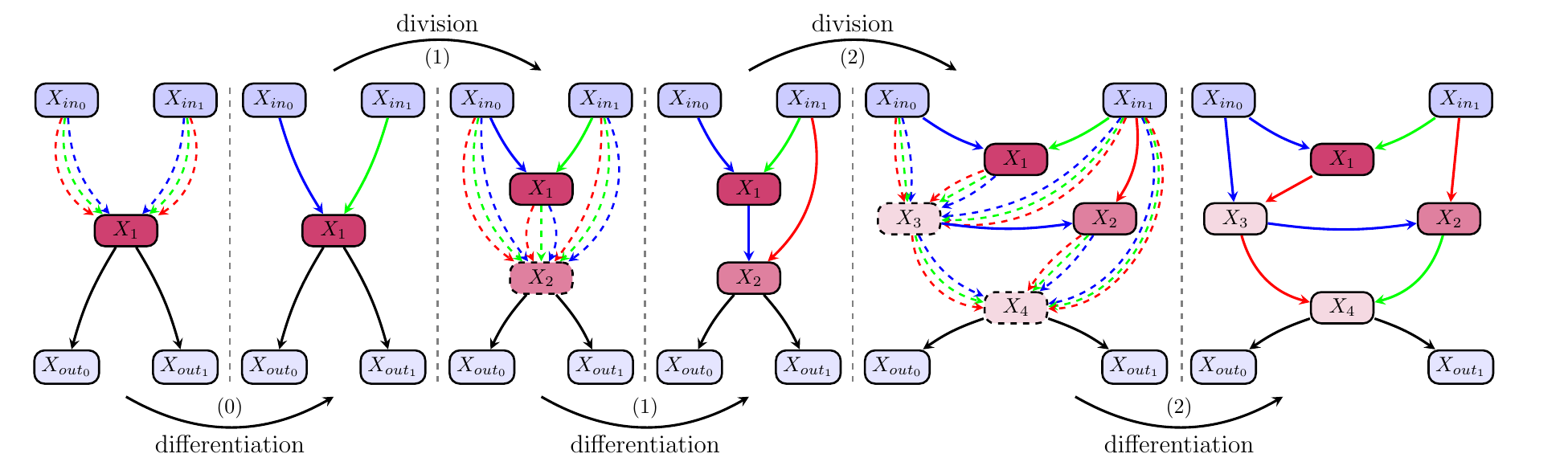}
    \vspace{-1.3em}
    \caption{
        \textbf{Illustration of network proliferation search paradigm.} 
        It comprises two procedures of network divison and differentiation. 
        The edges with different colors are associated with different operations. 
        A group of dashed edges denotes a mixture operation. 
        A local supernet comprises of three mixture operations pointing to one feature vertex. 
        (i) denotes the \emph{i-th} iteration. 
    }
    \vspace{-1.0em}
    \label{fission_figure}
\end{figure*}

\noindent
\textbf{Relation search space. }
The computation structure derived by relation search space is also a directed acyclic graph with the same number of vertices. 
Each vertex $\bm{E}_{i}$ is a latent relation representation (\ie, edge features in a GNN layer, as shown in the right of Figure \ref{space_figure}). 
The directed link $(i, j)$ in the DAG is associated with a relation-mining operation $\bm{o}_{\mathcal{E}}^{(i,j)}$. 
It extracts relational information from node features $\bm{V}_{i}$. 
The extracted information is joined with $\bm{E}_{i}$ to obtain higher order relation representation $\bm{E}_{i \rightarrow j} = \bm{o}_{\mathcal{E}}^{(i, j)}(\bm{V}_{i}, \bm{E}_{i}, h_{i,j})$. $h_{i,j}$ is a relation-mining function, such as \emph{substraction}, \emph{hardmard product}, \emph{gauss kernel}, \etc. 
Given a specific edge $(s,t)$ on the graph, $\bm{E}_{i \rightarrow j}^{s,t}$ can be computed using feature-wise linear modulation (FiLM): 
\begin{gather}
    \begin{split}
        \bm{E}_{i \rightarrow j}^{s,t} &= \bm{\gamma}_{s,t} \odot \bm{E}_{i}^{s,t} + \bm{\beta}_{s,t},  \\
        \bm{\gamma}_{s,t}, \bm{\beta}_{s,t} &= h_{i,j}(\bm{V}_{i}^{s}, \bm{V}_{i}^{t}; \bm{\theta}), 
    \end{split}
\end{gather}
where $\bm{\gamma}_{s,t}, \bm{\beta}_{s,t}$ is the affine transformation learned by $h_{i,j}$, $\bm{\theta}$ is the learnable parameters. 
This allows adaptive relational information fusion. 
Similar to node search space, we assume that each intermediate vertex has and only has two inputs, \ie, $\bm{E}_{i} = \bm{E}_{p_1 \rightarrow i} + \bm{E}_{p_2 \rightarrow i}$, where $p_1,p_2 \in \{in_0, in_1, 1, 2, \cdots, i-1\}$. 
Different relation-mining operations focus on extracting different types of relational information. 
For example, \emph{substraction} captures the relative change while \emph{hardmard product} emphasizes on the commonalities between the edge-connected nodes. 
The relation search space has 8 candidate relation-mining operations: \emph{E\_SUB}, \emph{E\_GAUSS}, \emph{E\_HAD}, \emph{E\_MAX}, \emph{E\_SUM}, \emph{E\_MEAN}, \emph{skip-connect}, and \emph{zero} operation. 
Eight candidate options of the relation-mining function are detailed in the {supplementary material}.



\subsection{Network Proliferation Search Paradigm}

\begin{algorithm}[t]
    \caption{Network Proliferation Search Paradigm}
    \label{algorithm}
    \hspace*{0.02in} {\bf Input:}
    a search algorithm $\mathcal{A}$, architecture size $\mathcal{S}$ \\
    \hspace*{0.02in} {\bf Output:}
    a graph neural architecture defined by $\{\mathbb{V},\mathbb{L}\}$ \\
    \hspace*{0.02in} {\bf Define:} $e(X_{s}, X_{t}, O)$ is a link from $X_s$ to $X_t$ with an operation $O$, where $O \in \{\bm{o}, \bm{\bar{o}}\}$, $\bm{\bar{o}}$ is the mixture operation
    \begin{algorithmic}[1]
    \State $\mathbb{V} \leftarrow \{X_1\}$ 
    \State $\mathbb{L} \leftarrow \{e(X_{in_0}, X_1, \bm{\bar{o}}), e(X_{in_1}, X_1, \bm{\bar{o}})\}$
    \While{True}
        \State \textbf{// network differentiation}
        \State $\mathbb{V}_{r} \leftarrow \mathbb{V} \cup \{X_{in_0}, X_{in_1}\}$
        \State Create a graph neural architecture $\mathcal{G}$ from $\{\mathbb{V}_{r}, \mathbb{L}\}$
        \State Initialize $\mathcal{G}$ with new parameters
        \State Perform search algorithm: $\mathbb{L} \leftarrow \mathcal{A}(\mathcal{G})$
        \If{$len(\mathbb{V}) \geqslant \mathcal{S}$ }
            \State \Return $\{\mathbb{V}_{r}, \mathbb{L}\}$
        \EndIf
        \State \textbf{// network division}
        \State $\mathbb{V}_{tmp} \leftarrow \mathbb{V}$, $\mathbb{L}_{tmp} \leftarrow \mathbb{L}$, $l \leftarrow len(\mathbb{V})$
        \For{$X_i$ in $\mathbb{V}_{tmp}$}
            \State $\mathbb{V} \leftarrow \mathbb{V} \cup \{X_{i+l}\}$
            \State $\mathbb{L} \leftarrow \mathbb{L} \cup \{e(X_{i}, X_{i + l}, \bm{\bar{o}})\}$
            \For{$e(X_s, X_t, O)|_{t = i}$ in $\mathbb{L}_{tmp}$}
                \State $\mathbb{L} \leftarrow \mathbb{L} \cup \{e(X_{s}, X_{i+l}, \bm{\bar{o}})\}$
            \EndFor
        \EndFor
        \State $\mathbb{L}_{tmp} \leftarrow \mathbb{L}$
        \For {$e(X_{s}, X_{t}, O)|_{s \in \mathbb{V}_{tmp}, s+l \neq t}$ in $\mathbb{L}_{tmp}$}
            \State $\mathbb{L} \leftarrow \mathbb{L} \cup \{e(X_{s+l}, X_{t}, O)\}  / \{e(X_{s}, X_{t}, O)\}$
        \EndFor
    \EndWhile
    \end{algorithmic}
\end{algorithm}



We detail how the proposed network proliferation search paradigm explores a single search space (\eg, node or relation search space) for simplicity. 
Since our dual relation-aware graph search space comprising node and relation search space is symmetric, it can be generalized to search the whole dual space simultaneously. 

Motivated by the observation that a well-performed small architecture can provide a good skeleton for building a larger one (shown in Figure \ref{box}), we can efficiently obtain the expected large architecture through iterative search based on the current small architecture. 
We detail this procedure in Figure \ref{fission_figure} and Algorithm \ref{algorithm}. 
Each iterative search comprises of two subprocedures, \ie, network division, and network differentiation. 
Network division builds a $2\times$ larger architecture based on the current architecture, where each feature vertex is divided into two parts. 
One part that retains original operations and connections is called parent vertex ($X_{p}$), whose role is to provide the architecture skeleton. 
The other one is called newly divided vertex ($X_{n}$), which joins $X_{p}$ and $X_{p}$'s two input vertices $X_{p_1}, X_{p_2}$ to build a local supernet. 
We define the mixture operation is parameterized by architectural parameters $\bm{\alpha}^{(i, j)}$ as a softmax mixture over all the possible operations within the operation space $\mathcal{O}$, \ie, $\bm{\bar{o}}^{(i, j)}(X_{i}) = \sum_{\bm{o} \in \mathcal{O}} \frac{exp(\alpha^{(i, j)}_{o})}{\sum_{o' \in \mathcal{O}} exp(\alpha^{(i, j)}_{o'})} \bm{o}(X_{i})$. 
The local supernet contains three mixture operations, which is represented as follows,
\begin{equation}
    X_{n} = \bm{\bar{o}}^{(p, n)}(X_{p}) + \bm{\bar{o}}^{(p_1, n)}(X_{p_1}) + \bm{\bar{o}}^{(p_2, n)}(X_{p_2}). 
\end{equation}
Since the local supernet contains $C^{2}_{3} \times |\mathcal{O}|$ candidate local subnets, we need to discretize it to one specific subnet. 
This is called network differentiation. 
During this procedure, we can adopt any search algorithms \cite{DARTS, SGASSG, Chen2019ProgressiveDA, He2020MiLeNASEN, Xie2019SNASSN, Xu2019PCDARTSPC}, that can discretize the supernet to the subnet. 
After the network differentiation, each mixture operation in the architecture is replaced with one specific operation. 
The algorithm terminates until the size of the architecture reaches the pre-defined threshold. 

\iftrue
\begin{figure}[t] 
    \centering
    \includegraphics[scale=0.60, trim = 0 0 0 0, clip]{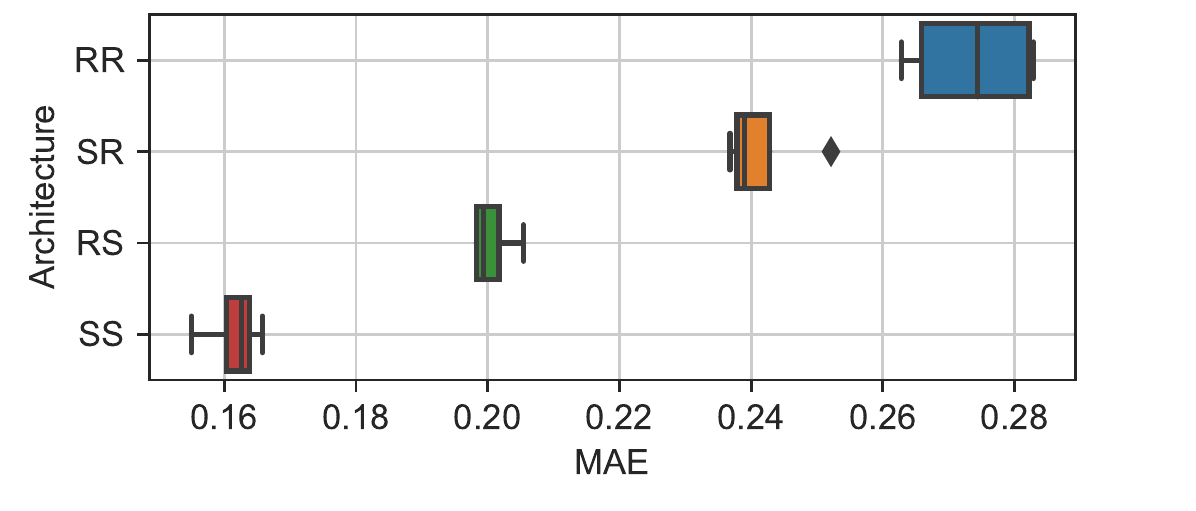}
    \vspace{-1em}
    \caption{
        \textbf{Boxplots of 8-layer architectures on ZINC. }
        ``R" and ``S" denote the random and SGAS \cite{SGAS} search strategies, respectively. 
        \eg, architecture ``RS" is constructed by sequentially performing random search, network division and SGAS search. 
        The architecture size after the first and the second search is 4 and 8. 
        Comparing ``RR'' with ``SR'', ``RS'' with ``SS'', we find that a pre-searched architecture is useful for subsequent search. 
    }
    \vspace{-1.0em}
    \label{box}
\end{figure}
\fi

Next, we analyze the complexity of our network proliferation search paradigm. 
The running time and occupied memory are proportional to the number of operations. 
It requires $log(N)$ iterations for searching for architecture with $N$ feature vertices. 
During the \emph{i-th} iteration, there are $2^{i-1} \times 2$ operations and $2^{i-1} \times 3$ mixture operations, \ie, $2^{i} + 3 \times 2^{i-1} |\mathcal{O}|$ operations in total. 
So the whole complexity is $O(\sum_{i=1}^{log(N)} 2^{i} + 3 \times 2^{i-1} |\mathcal{O}| ) = O(N|\mathcal{O}|)$. 
Compared with vanilla DARTS\cite{DARTS} ($O(N^{2}|\mathcal{O}|)$ complexity), our method reduces the complexity from quadratic to linear. 
This enables ours to directly search for large neural architectures without relying on the cell trick, and this also enlarges the global search space. 
For example, we try to search for an architecture with $N=16, |\mathcal{O}| = 8$. 
In general, due to the high complexity, the DARTS stacks $4$ cells with $4$ feature vertices in each of them, resulting in $3.0\times 10^9$ candidate architectures. 
Compared with vanilla DARTS, our search paradigm can explore up to $3.4 \times 10^{36}$ candidate architectures using comparable resources. 

\section{Experiments}

\subsection{Experimental Setups}


\begin{table*}[t]
    \centering
    \footnotesize
    \setlength{\tabcolsep}{0.7 mm}
    \caption{
        \textbf{Comparison with state-of-the-art architectures on the CLUSTER, ZINC, CIFAR10 and TSP datasets. }
        \textcircled{m} denotes the architecture is mannually designed. 
        The indicator \textbf{E} denotes whether the architecture can learn edge feature. 
        The ARGNP without edge feature means that the relation space is removed from relation-aware graph search space. 
        Note that mean and standard deviation are computed across 4 independently searched GNN architectures. 
    }
    \label{main}
    \begin{tabular}{@{}cccccccccccccc@{}}
    \toprule
                                                             &              & \multicolumn{3}{c}{\textbf{Node Level}}                             & \multicolumn{6}{c}{\textbf{Graph Level}}                                                                                           & \multicolumn{3}{c}{\textbf{Edge Level}}          \\ \cmidrule(l){3-5} \cmidrule(l){6-11} \cmidrule(l){12-14} 
    \multicolumn{1}{l}{\bf{Architecture}}                    &              & \multicolumn{3}{c}{\textbf{CLUSTER}}                                & \multicolumn{3}{c}{\textbf{ZINC}}                & \multicolumn{3}{c}{\textbf{CIFAR10}}                                            & \multicolumn{3}{c}{\textbf{TSP}}                 \\ \cmidrule(l){3-5} \cmidrule(l){6-8} \cmidrule(l){9-11} \cmidrule(l){12-14} 
                                                             & \textbf{E}   & \textbf{Metric}             & \textbf{Params}    & \textbf{Search}  & \textbf{Metric}               & \textbf{Params}  & \textbf{Search} & \textbf{Metric}                & \textbf{Params} & \textbf{Search}     & \textbf{Metric}         & \textbf{Params}   & \textbf{Search}       \\ 
                                                             & \CheckedBox  & \textbf{(AA \%) $\uparrow$} & \textbf{(M)}       & \textbf{(day)}   & \textbf{(MAE) $\downarrow$}   & \textbf{(M)}     & \textbf{(day)}  & \textbf{(OA \%) $\uparrow$}    & \textbf{(M)}    & \textbf{(day)}      & \textbf{(F1) $\uparrow$}& \textbf{(M)}      & \textbf{(day)}        \\ \midrule 
    \multicolumn{1}{l}{GCN~\cite{GCN}}                       & $\times$     & $68.50_{\pm0.98}$           &  $0.50$            & \textcircled{m}  & $0.367_{\pm0.011}$            & $0.50$           & \textcircled{m} & $56.34_{\pm0.38}$              & $0.10$          & \textcircled{m}     & $0.630_{\pm0.001}$      &  $0.10$           & \textcircled{m}       \\
    \multicolumn{1}{l}{GIN~\cite{GIN}}                       & $\times$     & $64.72_{\pm1.55}$           &  $0.52$            & \textcircled{m}  & $0.526_{\pm0.051}$            & $0.51$           & \textcircled{m} & $55.26_{\pm1.53}$              & $0.10$          & \textcircled{m}     & $0.656_{\pm0.003}$      &  $0.10$           & \textcircled{m}       \\
    \multicolumn{1}{l}{GraphSage~\cite{SAGE}}                & $\times$     & $63.84_{\pm0.11}$           &  $0.50$            & \textcircled{m}  & $0.398_{\pm0.002}$            & $0.51$           & \textcircled{m} & $65.77_{\pm0.31}$              & $0.10$          & \textcircled{m}     & $0.665_{\pm0.003}$      &  $0.10$           & \textcircled{m}       \\
    \multicolumn{1}{l}{GAT~\cite{GAT}}                       & $\times$     & $70.59_{\pm0.45}$           &  $0.53$            & \textcircled{m}  & $0.384_{\pm0.007}$            & $0.53$           & \textcircled{m} & $64.22_{\pm0.46}$              & $0.11$          & \textcircled{m}     & $0.671_{\pm0.002}$      &  $0.10$           & \textcircled{m}       \\
    \multicolumn{1}{l}{GatedGCN~\cite{GG}}                   & \checkmark   & $76.08_{\pm0.34}$           &  $0.50$            & \textcircled{m}  & $0.214_{\pm0.013}$            & $0.51$           & \textcircled{m} & $67.31_{\pm0.31}$              & $0.10$          & \textcircled{m}     & $0.838_{\pm0.002}$      &  $0.53$           & \textcircled{m}       \\
    \multicolumn{1}{l}{PNA~\cite{PNA}}                       & $\times$     & N/A                         &  N/A               & N/A              & $0.320_{\pm0.032}$            & $0.39$           & \textcircled{m} & $70.46_{\pm0.44}$              & $0.11$          & \textcircled{m}     & N/A                     &  N/A              & N/A                   \\
    \multicolumn{1}{l}{PNA~\cite{PNA}}                       & \checkmark   & N/A                         &  N/A               & N/A              & $0.188_{\pm0.004}$            & $0.39$           & \textcircled{m} & $70.47_{\pm0.72}$              & $0.11$          & \textcircled{m}     & N/A                     &  N/A              & N/A                   \\
    \multicolumn{1}{l}{DGN~\cite{DGN}}                       & $\times$     & N/A                         &  N/A               & N/A              & $0.219_{\pm0.010}$            & $0.39$           & \textcircled{m} & $72.70_{\pm0.54}$              & $0.11$          & \textcircled{m}     & N/A                     &  N/A              & N/A                   \\
    \multicolumn{1}{l}{DGN~\cite{DGN}}                       & \checkmark   & N/A                         &  N/A               & N/A              & $0.168_{\pm0.003}$            & $0.39$           & \textcircled{m} & $72.84_{\pm0.42}$              & $0.11$          & \textcircled{m}     & N/A                     &  N/A              & N/A                   \\
    \multicolumn{1}{l}{GNAS-MP~\cite{Cai2021RethinkingGN}}   & $\times$     & $74.77_{\pm0.15}$           &  $1.61$            & $0.80$           & $0.242_{\pm0.005}$            & $1.20$           & $0.40$          & $70.10_{\pm0.44}$              & $0.43$          & $3.20$              & $0.742_{\pm0.002}$      &  $1.20$           & $2.10$                \\ \midrule 
    \multicolumn{1}{l}{ARGNP (2)}                            & $\times$     & $61.61_{\pm0.27}$           &  $0.07$            & $0.04$           & $0.430_{\pm0.003}$            & $0.09$           & $0.01$          & $66.55_{\pm0.13}$              & $0.10$          & $0.11$              & $0.655_{\pm0.003}$      &  $0.09$           & $0.05$                \\ 
    \multicolumn{1}{l}{ARGNP (4)}                            & $\times$     & $64.06_{\pm0.45}$           &  $0.14$            & $0.07$           & $0.303_{\pm0.013}$            & $0.14$           & $0.01$          & $66.65_{\pm0.39}$              & $0.18$          & $0.14$              & $0.668_{\pm0.003}$      &  $0.17$           & $0.06$                \\ 
    \multicolumn{1}{l}{ARGNP (8)}                            & $\times$     & $68.73_{\pm0.12}$           &  $0.25$            & $0.20$           & $0.239_{\pm0.009}$            & $0.27$           & $0.02$          & $67.37_{\pm0.32}$              & $0.33$          & $0.48$              & $0.674_{\pm0.002}$      &  $0.29$           & $0.21$                \\ 
    \multicolumn{1}{l}{ARGNP (16)}                           & $\times$     & $71.92_{\pm0.29}$           &  $0.53$            & $0.71$           & $0.221_{\pm0.004}$            & $0.51$           & $0.06$          & $67.10_{\pm0.51}$              & $0.58$          & $1.77$              & $0.684_{\pm0.002}$      &  $0.56$           & $0.76$                \\ \midrule
    \multicolumn{1}{l}{ARGNP (2)}                            & \checkmark   & $64.99_{\pm0.31}$           &  $0.08$            & $0.06$           & $0.318_{\pm0.009}$            & $0.08$           & $0.01$          & $69.14_{\pm0.30}$              & $0.10$          & $0.17$              & $0.773_{\pm0.001}$      &  $0.08$           & $0.08$                \\ 
    \multicolumn{1}{l}{ARGNP (4)}                            & \checkmark   & $74.75_{\pm0.25}$           &  $0.15$            & $0.09$           & $0.197_{\pm0.006}$            & $0.15$           & $0.01$          & $71.83_{\pm0.32}$              & $0.17$          & $0.23$              & $0.821_{\pm0.001}$      &  $0.14$           & $0.10$                \\ 
    \multicolumn{1}{l}{ARGNP (8)}                            & \checkmark   & $76.32_{\pm0.03}$           &  $0.29$            & $0.31$           & $0.155_{\pm0.003}$            & $0.28$           & $0.04$          & $73.72_{\pm0.32}$              & $0.33$          & $0.84$              & $0.841_{\pm0.001}$      &  $0.30$           & $0.39$                \\ 
    \multicolumn{1}{l}{ARGNP (16)}                           & \checkmark   & \textcolor{red}{$\bm{77.35_{\pm0.05}}$}  &  $0.52$            & $1.10$           & \textcolor{red}{$\bm{0.136_{\pm0.002}}$}   & $0.52$           & $0.15$          & \textcolor{red}{$\bm{73.90_{\pm0.15}}$}     & $0.64$          & $2.95$              & \textcolor{red}{$\bm{0.855_{\pm0.001}}$}   &  $0.62$                & $1.23$                     \\ 
    \bottomrule 
    \end{tabular}
    \vspace{-1em}
\end{table*}

\noindent
\textbf{Dataset. }
We evaluate our method on six datasets, \ie, ZINC~\cite{ZINC}, CLUSTER~\cite{bench}, CIFAR10~\cite{CIFAR10}, TSP~\cite{bench}, ModelNet10~\cite{ModelNet}, ModelNet40~\cite{ModelNet} across four different graph learning tasks (node classification, edge classification, graph classification and graph regression). 
ZINC is one popular real-world molecular dataset of 250K graphs, whose task is graph property regression, out of which we select 12K for efficiency following works~\cite{Cai2021RethinkingGN, PNA, bench}. 
CLUSTER is the node classification tasks generated via Stochastic Block Models~\cite{SBM}, which are used to model communications in social networks by modulating the intra-community and extra-community connections. 
CIFAR10 is the original classical image classification dataset and converted into graphs using superpixel~\cite{SLIC} algorithm to test graph classification task. 
TSP dataset is based on the classical \emph{Travelling Salesman Problem}, which tests edge classification on 2D Euclidean graphs to identify edges belonging to the optimal TSP solution. 
ModelNet is a dataset for 3D object recognition with two variants, ModelNet10 and ModelNet40, which comprise objects from 10 and 40 classes, respectively. 
We sample 1024 points for each object as input and use \emph{k-NN} algorithm to construct the edges ($k = 9$ by default unless it is specified). 

\noindent
\textbf{Searching settings. }
The original architecture is initialized with 2 feature vertices. 
We perform \emph{network proliferation} for 4 iterations to obtain a sequence of GNN architectures with the size of $\{2,4,8,16\}$. 
Specifically, we choose SGAS~\cite{SGAS} as the search strategy that can differentiate the local supernet to a specific subnet. 
During the network differentiation, after warming up for 10 epochs, SGAS begins to simultaneously determine one node-learning operation and one relation-mining operation for every 5 epochs. 
Thus, the search epoch is set to $\{25, 25, 45, 85\}$ for 4 sequential iterations. 
To carry out the architecture search, we hold out half of the training data as the validation set. 
For one-shot differentible search strategies (SGAS~\cite{SGAS} and DARTS~\cite{DARTS}), there are operation weights $\bm{w}$ and architectural parameters $\bm{\alpha}$ to be optimized. 
We use momentum SGD to optimize the weights $\bm{w}$, with initial learning rate $\eta_{\bm{w}}=0.025$ (anneald down to zero following a cosine schedule without restart), momentum $0.9$, and weight decay $3 \times 10^{-4}$. 
We use Adam~\cite{ADAM} as the optimizer for $\bm{\alpha}$, with initial learning rate $\eta_{\alpha} = 3 \times 10^{-4}$, momentum $\beta = (0.5, 0.999)$ and weight decay $10^{-3}$.

\noindent
\textbf{Training settings. }
We follow all the training settings~(data splits, optimizer, metrics, \etc) in work~\cite{Cai2021RethinkingGN,bench}. 
Specifically, we adopt Adam~\cite{ADAM} with the same learning rate decay for all runs. 
The learning rate is initialized with $10^{-3}$, which is reduced by half if the validation loss stops decreasing after $20$ epochs. 
The weight decay is set to $0$. 
The dropout is set to $0.5$ to alleviate the overfitting. 
Our architectures are all trained for $400$ epochs with a batch size of $32$. 
We report the mean and standard deviation of the metric on the test dataset of $4$ discovered architectures. 
These experiments are run on a single NVIDIA GeForce RTX 3090 GPU.

\subsection{Results and Analysis}

In Table~\ref{main} and Table~\ref{modelnet}, we compare our ARGNP with the state-of-the-art hand-crafted and search-based GNN architectures on the CLUSTER, ZINC, CIFAR10, TSP, ModelNet10, and ModelNet40 datasets. 
The evaluation metric is the average accuracy (AA) for CLUSTER, mean absolute error (MAE) for ZINC, F1-score (F1) for TSP. 
For CIFAR10, ModelNet10, and ModelNet40, we use the overall accuracy (OA) as the evaluation metric. 
To make a fair comparison, we also report the architecture parameters, the search cost, and the mean and standard deviation of all the metrics. 
We can see that, on all the six datasets for four classical graph learning tasks, the GNN architectures discovered by our ARGNP surpass the state-of-the-art architectures by a large margin in terms of both mean and standard deviation. 
Compared with the state-of-the-art search-based method GNAS-MP \cite{Cai2021RethinkingGN}, our searched architecture can easily achieve better performance with only $\frac{1}{10}\sim\frac{1}{4}$ parameters. 
This benefits from that the relation-aware graph search space can mine hierarchical relation information (such as local structural similarity) to guide anisotropic message passing. 
Moreover, the network proliferation search paradigm can efficiently and effectively explore the proposed search space. 
We visualize the best-performed GNN architecture with the size of 4 in Figure \ref{ZINC_cell}, which is searched on the ModelNet40 dataset. 
Other examples are provided in the supplementary material.

\begin{figure*}[t]

    \centering
    \includegraphics[scale = 0.55, trim = 30 30 30 30, clip]{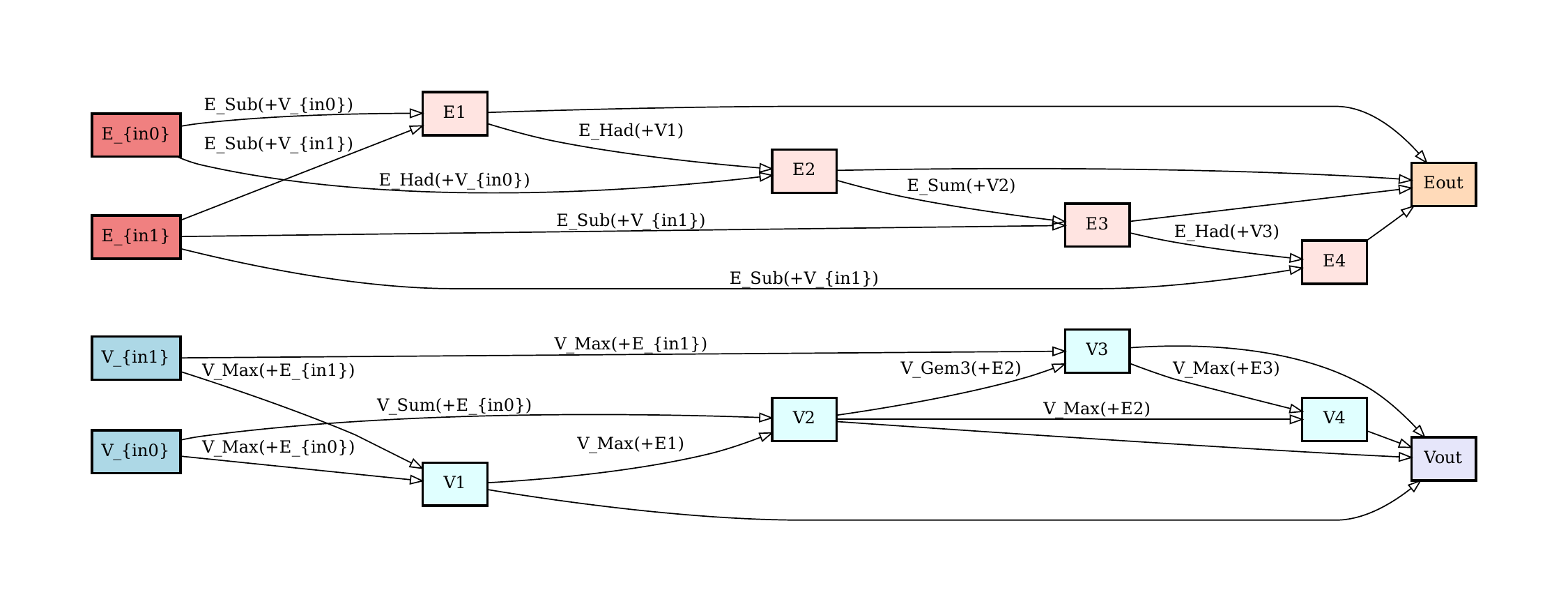}
    \vspace{-0.5em}
    \caption{
        \textbf{The best GNN architecture with the network size of 4 searched on the ModelNet40 dataset. }
    }
    \vspace{-1em}
    \label{ZINC_cell}

\end{figure*}

\subsection{Ablation of Search Space}

We study the influence of the relation search space in our proposed relation-aware graph search. 
First, we construct a search space variant by removing the relation search space. 
Then we perform GNN architecture search on this variant using the network proliferation search paradigm and obtain a sequence of GNN architectures with the size of $\{2,4,8,16\}$. 
These GNN architectures are evaluated on six datasets. 
For a fair comparison, we increase the dimension of the node features to keep the architectural parameters comparable. 
As shown in Table \ref{main} and Table \ref{modelnet}, the best performance of the search space variant without relation learning descends by a large margin. 
Under all the different network size settings, relation learning can significantly improve the capability of graph reasoning. 
Interestingly, this improvement is also observed on the CLUSTER, CIFAR10, and ModelNet datasets which don't have original edge features. 
Taking the CLUSTER dataset as an example, it aims at identifying the community clusters, where the graphs represent the community networks. 
The edges play a role in connecting two nodes and have no original meaningful features. 
In this case, relation learning can mine hierarchical relational information by extracting local structural similarities between nodes. 
This can help distinguish between intra-community and extra-community connections for learning better discriminative node features. 

\begin{table}[t]
    \centering
    \footnotesize
    \setlength{\tabcolsep}{0.5 mm}
    \caption{
        \textbf{Comparision with state-of-the-art architectures on the ModelNet10 and ModelNet40 datasets at 3D point cloud recognition task. }
        \textbf{L} denotes the size of GNN architecture. 
    }
    \label{modelnet}
    \begin{tabular}{@{}cccccccc@{}}
    \toprule
                                                        &                       &                   & \textbf{ModelNet10}        &   & \textbf{ModelNet40}           &                       &                       \\ \cmidrule(l){4-4} \cmidrule(l){6-6}
    \textbf{Architecture}                               & \textbf{L}            & \textbf{E}        & \textbf{Metric}            &   & \textbf{Metric}               & \textbf{Params}       & \textbf{Search}       \\
                                                        & \textbf{(\#)}         & \CheckedBox       & \textbf{(OA \%) $\uparrow$}&   & \textbf{(OA \%) $\uparrow$}   & \textbf{(M)}          & \textbf{(Day)}        \\ \midrule
    \multicolumn{1}{l}{3DmFV~\cite{3DmFV}}              & /                     &  /                & $95.2$                     &   & $91.6$                        & $45.77$               & \textcircled{m}       \\
    \multicolumn{1}{l}{PointNet++~\cite{PointNetA}}     & /                     &  /                & N/A                        &   & $90.7$                        & $1.48$                & \textcircled{m}       \\
    \multicolumn{1}{l}{PCNN~\cite{PCNN}}                & /                     &  /                & N/A                        &   & $92.3$                        & $8.20$                & \textcircled{m}       \\
    \multicolumn{1}{l}{PointCNN~\cite{PointCNN}}        & /                     &  /                & N/A                        &   & $92.2$                        & $0.60$                & \textcircled{m}       \\
    \multicolumn{1}{l}{DGCNN~\cite{DGCNN}}              & /                     &  /                & N/A                        &   & $92.2$                        & $1.84$                & \textcircled{m}       \\
    \multicolumn{1}{l}{KPConv~\cite{KPConv}}            & /                     &  /                & N/A                        &   & $92.9$                        & $14.3$                & \textcircled{m}       \\
    \multicolumn{1}{l}{SGAS~\cite{SGAS}}                & $9$                   & \checkmark        & N/A                        &   & $92.93_{\pm 0.19}$            & $8.87$                & $0.19$                \\ \midrule
    \multicolumn{1}{l}{ARGNP}                           & $2$                   & $\times$          & $93.20_{\pm 0.24}$         &   & $91.11_{\pm 0.24}$            & $1.80$                & $0.03$                \\ 
    \multicolumn{1}{l}{ARGNP}                           & $4$                   & $\times$          & $93.86_{\pm 0.25}$         &   & $91.30_{\pm 0.22}$            & $2.27$                & $0.04$                \\ 
    \multicolumn{1}{l}{ARGNP}                           & $8$                   & $\times$          & $94.23_{\pm 0.22}$         &   & $91.85_{\pm 0.18}$            & $3.20$                & $0.15$                \\ \midrule
    \multicolumn{1}{l}{ARGNP}                           & $2$                   & \checkmark        & $95.07_{\pm 0.31}$         &   & $92.47_{\pm0.23}$             & $2.50$                & $0.04$                \\ 
    \multicolumn{1}{l}{ARGNP}                           & $4$                   & \checkmark        & $95.35_{\pm 0.23}$         &   & $92.80_{\pm0.19}$             & $3.05$                & $0.06$                \\ 
    \multicolumn{1}{l}{ARGNP}                           & $8$                   & \checkmark        & \textcolor{red}{$\bm{95.87_{\pm0.22}}$}     &   & \textcolor{red}{$\bm{93.33_{\pm0.15}}$}        & $4.15$                & $0.20$                      \\ 
    \bottomrule
    \end{tabular}
    \vspace{-1.0em}
\end{table}

\subsection{Ablation of Search Paradigm}

To investigate the effectiveness of our \emph{Network Proliferation Search Paradigm (NPSP)}, we conduct the ablation experiments on ZINC dataset around \emph{network size}, \emph{search strategy}, \emph{whether to use cell trick} and \emph{whether to use NPSP}. 
We run 14 different experiments and report the results in Table \ref{proliferation}. 
We observe the following phenomena. 
First, \emph{the cell trick improves the search efficiency but weakens the expressive capability of graph search space}.
This results from its original assumption where the GNN architecture is a stack of the same building cells that narrows our relation-aware graph search space. 
Therefore, the search strategy with the cell trick performs worse than that without it, which is demonstrated by the contrast between exp 2 and exp 3, exp 5 and exp 6. 
Second, our \emph{NPSP can both significantly improves the search efficiency and search effect with different search strategies}. 
The performance improvement benefits from that the \emph{NPSP} can alleviate the subnet interference and mitigate the shrink of search space by breaking away from the cell assumption. 
The efficiency improvement lies in that NPSP shifts the training object from global supernet to sequential local supernets. 
They are supported by exp 4, 7, 11, and 14, where NPSP achieves the best performance with less time cost under all the experimental settings.




\begin{table}[t]
    \centering
    \footnotesize
    \setlength{\tabcolsep}{0.6 mm}
    \caption{
        \textbf{Performance of the relation-aware graph search space under different  settings. }
        \textbf{Cell} is an indicator of whether to use the cell trick. 
        \textbf{NPSP} is an indicator of whether to use the network proliferation search. 
        OOM denotes out of memory. 
    }
    \label{proliferation}
    \begin{tabular}{@{}ccccccccc@{}}
    \toprule
                          &                                                         & \multicolumn{7}{c}{\textbf{ZINC}}                                                                                                                                                                      \\ \cmidrule(l){3-9} 
     \textbf{\#}          & \textbf{Method}                                         & \textbf{L}                          & \textbf{Search}          & \textbf{Cell}  & \textbf{NPSP}                    & \textbf{Metric}                                 & \textbf{Params}               & \textbf{Search}                                    \\
                          &                                                         & \textbf{(\#)}                       & \textbf{Strategy}        & \CheckedBox    & \CheckedBox                      & \textbf{(MAE) $\downarrow$}                     & \textbf{(M)}                  & \textbf{(Day)}                                     \\ \midrule
       1                  & \multicolumn{1}{l}{R-space}                             & $8$                                 &  Random                  &  $\times$      &  $\times$                        & $0.303_{\pm0.058}$                              & $0.27$                        &  $0.$                                            \\
       2                  & \multicolumn{1}{l}{R-space}                             & $8$                                 &  DARTS                   &  \checkmark    &  $\times$                        & $0.160_{\pm0.005}$                              & $0.28$                        &  $0.17$                                            \\
       3                  & \multicolumn{1}{l}{R-space}                             & $8$                                 &  DARTS                   &  $\times$      &  $\times$                        & $0.157_{\pm0.008}$                              & $0.28$                        &  $0.30$                                            \\
       4                  & \multicolumn{1}{l}{R-space}                             & $8$                                 &  DARTS                   &  $\times$      &  \checkmark                      & $\bm{0.150_{\pm0.006}}$                         & $0.29$                        &  $\bm{0.08}$                                       \\
       5                  & \multicolumn{1}{l}{R-space}                             & $8$                                 &  SGAS                    &  \checkmark    &  $\times$                        & $0.165_{\pm0.008}$                              & $0.30$                        &  $0.13$                                            \\
       6                  & \multicolumn{1}{l}{R-space}                             & $8$                                 &  SGAS                    &  $\times$      &  $\times$                        & $0.161_{\pm0.008}$                              & $0.30$                        &  $0.25$                                            \\
       7                  & \multicolumn{1}{l}{R-space}                             & $8$                                 &  SGAS                    &  $\times$      &  \checkmark                      & $\bm{0.155_{\pm0.003}}$                         & $0.28$                        &  $\bm{0.06}$                                       \\ \cmidrule{2-9}
       8                  & \multicolumn{1}{l}{R-space}                             & $16$                                &  Random                  &  $\times$      &  $\times$                        & $0.185_{\pm0.024}$                              & $0.51$                        &  $0.$                                            \\
       9                  & \multicolumn{1}{l}{R-space}                             & $16$                                &  DARTS                   &  \checkmark    &  $\times$                        & $0.144_{\pm0.004}$                              & $0.57$                        &  $0.38$                                            \\
       10                 & \multicolumn{1}{l}{R-space}                             & $16$                                &  DARTS                   &  $\times$      &  $\times$                        &  N/A                                            &  N/A                          &  OOM                                               \\
       11                 & \multicolumn{1}{l}{R-space}                             & $16$                                &  DARTS                   &  $\times$      &  \checkmark                      & $\bm{0.139_{\pm0.005}}$                         & $0.56$                        &  $\bm{0.24}$                                       \\
       12                 & \multicolumn{1}{l}{R-space}                             & $16$                                &  SGAS                    &  \checkmark    &  $\times$                        & $0.140_{\pm0.003}$                              & $0.60$                        &  $0.32$                                            \\ 
       13                 & \multicolumn{1}{l}{R-space}                             & $16$                                &  SGAS                    &  $\times$      &  $\times$                        &  N/A                                            &  N/A                          &  OOM                                               \\
       14                 & \multicolumn{1}{l}{R-space}                             & $16$                                &  SGAS                    &  $\times$      &  \checkmark                      & $\bm{0.136_{\pm0.002}}$        & $0.52$                        &  $\bm{0.21}$                                       \\
    \bottomrule
    \end{tabular}
    \vspace{-1.0em}
\end{table}

\begin{figure*}[t]

    \centering
    \includegraphics[scale = 0.68, trim = 0 0 0 0, clip]{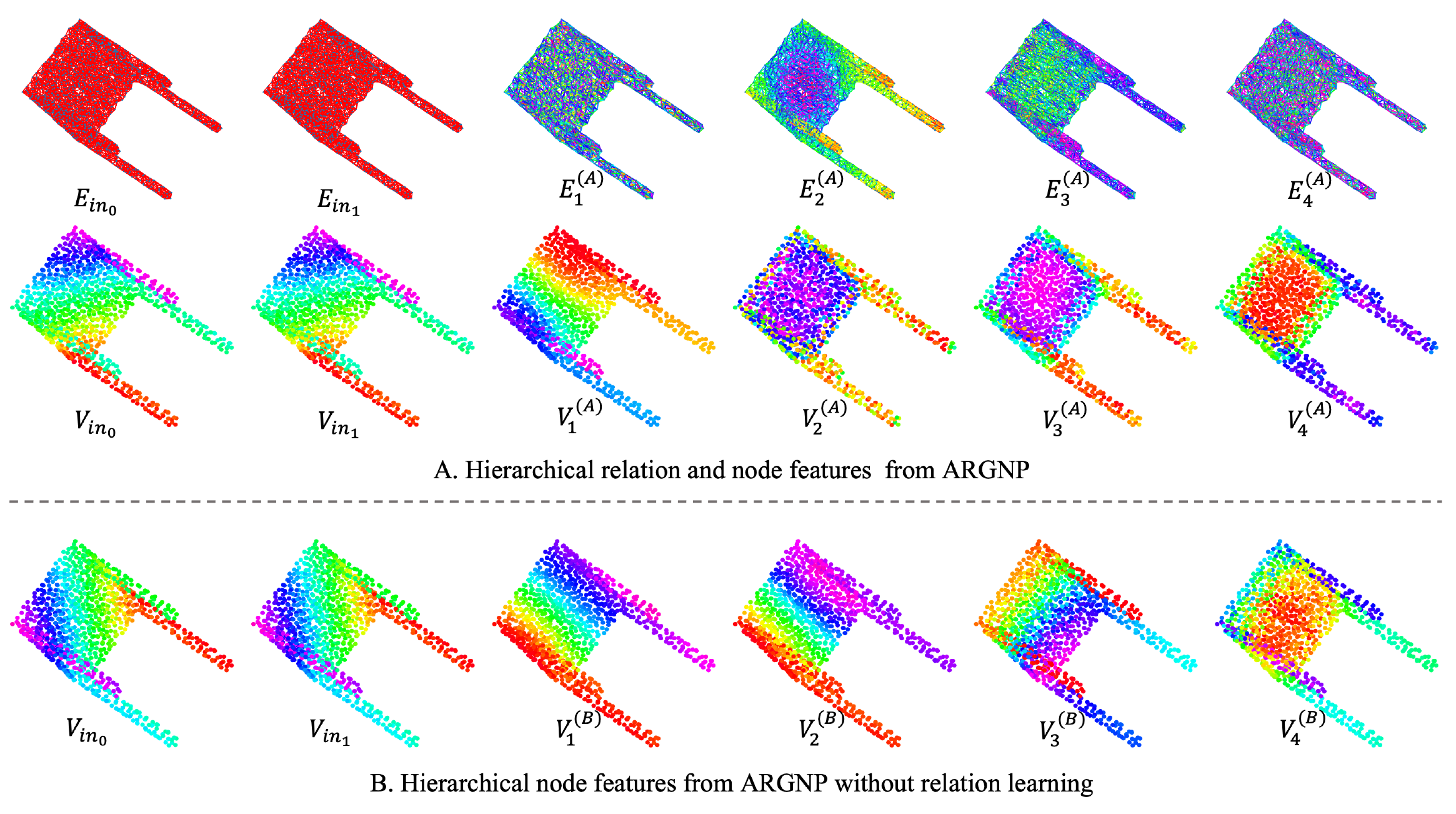}
    \caption{
        \textbf{Visualization of the learned hierarchical features for 3D point cloud recognition} (taking table as an example). 
        Relation features with different edge color distribution have different message passing preferences. 
        Node features with different node color distribution represent different clustering effects. 
    }
    \vspace{-1.0em}
    \label{pointcloud}
\end{figure*}

\subsection{Visualizing Hierarchical Features}

To better demonstrate the effectiveness of the relation learning, we provide relation and node features visualization on ModelNet40 dataset. 
During the inference, we feed forward one 3D pointcloud object into the network with the input $\{\bm{V}_{in_0}, \bm{V}_{in_1}, \bm{E}_{in_0}, \bm{E}_{in_1}\}$, where $\bm{V}_{in_0} = \bm{V}_{in_1} \in \mathbb{R}^{1024 \times 3}$ are the 3D coordinates and $\bm{E}_{in_0} = \bm{E}_{in_1} = \bm{1} \in \mathbb{R}^{(1024 \times 9)\times 1}$ are the pseudo relation features. 
The hierarchical node/relation features generated from each layer is denoted as $\{\bm{V}_{1}, \cdots, \bm{V}_{4}\}$ and $\{\bm{E}_1, \cdots, \bm{E}_{4}\}$, respectively. 
For better visualization on the point cloud graph, we reduce the feature dimension to 1 through principal component analysis (PCA).
The edges with a similar color are considered to have the same message passing preferences, while the nodes with a similar color are considered to belong to a similar cluster. 
A visualization example from ARGNP and a version without relation learning architecture are shown in Figure \ref{pointcloud}A and \ref{pointcloud}B, respectively. 

As shown in Figure \ref{pointcloud}A, ARGNP can capture the structural information and well discriminate different parts of the object (\eg, the legs, desktop, and border of the table in $V_4^{(A)}$). 
In contrast, as shown in Figure \ref{pointcloud}B, the original GNN without relation learning architecture can only gradually propagate the node information through the input graph based on 3D coordinates. 
As a result, the similar nodes that are distant in the 3D space can not be well clustered (\eg, 4 legs of the table in $V_4^{(B)}$). 
The above comparison shows that the relation features can guide better message passing mechanisms to learn more effective node features. 
To show the role of the relation learning more specifically, accompanied with the searched GNN in Figure \ref{ZINC_cell}, we analyze the features in Figure \ref{pointcloud}A. 
For example, the $E_1^{(A)}$ is learned by \emph{substraction} operation. The \emph{substraction} operation is similar to an ``border detection" operation that can discriminate the directions between two nodes. 
In this way, the ARGNP can directly distinguish groups of parallel components of an object (\eg, legs, parallel borders, \etc in $V_2^{(A)}$). 
However, without using the relation features, it requires numerous rounds of message passing which may cause the over smoothing problem. 
With the help of relation learning, the components with similar structures can be well clustered.

\section{Conclusion}
This paper proposes the automatic relation-aware graph network proliferation (ARGNP) method to design the optimal GNN architectures by devising a novel relation-aware graph search space and a network proliferation search paradigm. 
The search space significantly improves the upper bound of the discovered GNN's reasoning capability while the proliferation search paradigm promotes search efficiency and effectiveness. 
Experiments show that ARGNP achieves superior performance for four tasks on six datasets.


\textbf{Acknowledgement.} This work was supported in part by the National Key R\&D Program of China under Grant 2018AAA0102000, and in part by the National Natural Science Foundation of China: U21B2038, 61931008, 61732007, and CAAI-Huawei MindSpore Open Fund, Youth Innovation Promotion Association of CAS under Grant 2020108, CCF-Baidu Open Fund.


\clearpage
{\small
\bibliographystyle{ieee_fullname}
\bibliography{egbib}
}

\clearpage

\appendix

\section{Search Space Details}

In this section, we first detail how node-learning operations and relation-learning operations are formulated. 
Then we introduce the design of task-based layers for different graph learning tasks. 
$\bm{V} \in \mathbb{R}^{n \times d_V}$, $\bm{E} \in \mathbb{R}^{m \times d_E}$ are node features and relation features that are fed forward to node-learning and relation-learning operations, respectively. $n$ and $m$ are the number of nodes and edges of the input graph data. 

\subsection{Node-learning Operations}
As discussed in the body content, a node-learning operation $\bm{o}_{\mathcal{V}}^{(i, j)}$ computes the transformed information from feature vertex $\bm{V}_{i}$ to $\bm{V}_{j}$, which is denoted as $\bm{V}_{i \rightarrow j} = \bm{o}_{\mathcal{V}}^{(i, j)}(\bm{V}_{i}, \bm{E}_{i}, f_{i,j})$. 
Specifically, given node t on the graph, its transformed node feature $\bm{V}_{i \rightarrow j}^{t}$ is formulated as 
\begin{gather}
    \begin{split}
    \bm{V}_{i \rightarrow j}^{t} = f_{i,j}(\{\bm{\gamma}_{s,t} \odot &\bm{V}_{i}^{s} + \bm{\beta}_{s,t} | s \in \mathcal{N}(t)\}), \\
    \bm{\gamma}_{s,t}, \bm{\beta}_{s,t} &= g(\bm{E}_{i}^{s,t}; \bm{\theta}), 
    \end{split}
\end{gather}
where, $\mathcal{N}(t)$ denotes the set of neighbors of target node t on the graph, $f_{i,j}$ is an aggregating function. 
$g(\cdot)$ is a two-layer multilayer perceptron to compute the affine transformation with the learnable parameters $\bm{\theta} = \begin{bmatrix}\bm{W^{1}}& \bm{W^{2}}& \bm{W^{k}}& \bm{W^{b}}\end{bmatrix}$, which is formulated as 
\begin{equation}
    \begin{bmatrix}\bm{\gamma}_{s,t} & \bm{\beta}_{s,t}\end{bmatrix} = \sigma(\sigma(\bm{E}_{i}^{s,t} \bm{W^{1}}) \bm{W^{2}})\begin{bmatrix}\bm{W^{k}} & \bm{W^{b}}\end{bmatrix},
\end{equation}
where $\sigma(\cdot)$ is the rectified linear unit. 
The difference between different node-learning operations lies in the choice of message aggregating functions. 
We design 8 candidate aggregating function options, \ie, \emph{V\_MEAN}, \emph{V\_SUM}, \emph{V\_MAX}, \emph{V\_STD}, \emph{V\_GEM2}, \emph{V\_GEM3}, \emph{zero} and \emph{skip-connect} for capturing different types of information. 
For clarity, we denote $\bm{M}_{s,t}$ as the modulated incoming message from node t to node s, where $\bm{M}_{s,t} = \bm{\gamma}_{s,t} \odot \bm{V}_{i}^{s} + \bm{\beta}_{s,t}$. 
The mean aggregation of neighboring messages is denoted as 
\begin{equation}
\mu_{t}(\bm{M}) = \frac{1}{|\mathcal{N}(t)|}\sum\limits_{s \in \mathcal{N}(t)} \bm{M}_{s,t}. 
\end{equation}

\noindent
\emph{\textbf{V\_MEAN}} is the average of neighboring messages, which captures the mean statistics of neighboring messages \cite{SAGE}, written as 
\begin{equation}
    \bm{V}_{i \rightarrow j}^{t} = \mu_{t}(\bm{M}).
\end{equation}

\noindent
\emph{\textbf{V\_SUM}} is the sum of neighboring messages, which captures local structural information \cite{GIN}, written as
\begin{equation}
    \bm{V}_{i \rightarrow j}^{t} = |\mathcal{N}(t)| \times \mu_{t}(\bm{M}). 
\end{equation}

\noindent
\emph{\textbf{V\_MAX}} is the max of neighboring messages, which captures the representative information \cite{GIN}, written as
\begin{equation}
    \bm{V}_{i \rightarrow j}^{t} = \max\limits_{s \in \mathcal{N}(t)} \bm{M}_{s,t}.
\end{equation}

\noindent
\emph{\textbf{V\_STD}} is the standard deviation of input feature set, which captures the stability of neighboring messages \cite{PNA}, \ie, 
\begin{equation}
    \bm{V}_{i \rightarrow j}^{t} = \sqrt{ReLU(\mu_{t}(\bm{M}^2) - \mu_{t}(\bm{M})) + \epsilon},
\end{equation}
where \emph{ReLU} is the rectified linear unit used to avoid negative values caused by numerical errors and $\epsilon$ is a small positive number to ensure the output is differentiable \cite{PNA}. 

Moreover, we also introduce Generalized Mean Pooling (GeM) for aggregating messages, which can focus on learning to propagate the prominent message \cite{GEM}, written as 
\begin{equation}
    GeM_{t}(\bm{M}, \alpha) = \sqrt[\alpha]{ReLU(\mu_{t}(\bm{M}^{\alpha})) + \epsilon}, 
\end{equation}
where $\alpha$ is the hyper parameter. 
Here, we adopt two widely used parameters to construct the node-learning operations, \ie, \emph{V\_GEM2} and \emph{V\_GEM3}. 

\noindent
\emph{\textbf{V\_GEM2}}:
\begin{equation}
    \bm{V}_{i \rightarrow j}^{t} = GeM_{t}(\bm{M}, 2).
\end{equation}

\noindent
\emph{\textbf{V\_GEM3}}:
\begin{equation}
    \bm{V}_{i \rightarrow j}^{t} = GeM_{t}(\bm{M}, 3).
\end{equation}

\noindent
\emph{\textbf{skip-connect}} is to enhance the central node information and mitigate the gradient vanishing, written as 
\begin{equation}
    \bm{V}_{i \rightarrow j}^{t} = \bm{V}_{i}^{t}.
\end{equation}

\noindent 
\emph{\textbf{zero}} operation is included in the search space to indicate a lack of connection. 
Links that are important should have a low weight in the \emph{zero} operation \cite{SGAS}. 
It is formulated as
\begin{equation}
    \bm{V}_{i \rightarrow j}^{t} = \bm{0.} \times  \bm{V}_{i}^{t}.
\end{equation}

\subsection{Relation-mining Operations}

A relation-mining operation $\bm{o}_{\mathcal{E}}^{(i,j)}$ computes the transformed relational information $\bm{E}_{i \rightarrow j} = \bm{o}_{\mathcal{E}}^{(i, j)}(\bm{V}_{i}, \bm{E}_{i}, h_{i,j})$. $h_{i,j}$ is a relation-mining network, such as \emph{substraction}, \emph{hardmard product}, \emph{gauss kernel}, \etc. 
Specifically, given a specific edge $(s,t)$ on the graph, $\bm{E}_{i \rightarrow j}^{s,t}$ is computed using feature-wise linear modulation: 
\begin{gather}
    \begin{split}
        \bm{E}_{i \rightarrow j}^{s,t} &= \bm{\gamma}_{s,t} \odot \bm{E}_{i}^{s,t} + \bm{\beta}_{s,t},  \\
        \bm{\gamma}_{s,t}, \bm{\beta}_{s,t} &= h_{i,j}(\bm{V}_{i}^{s}, \bm{V}_{i}^{t}; \bm{\theta}), 
    \end{split}
\end{gather}
$\bm{\gamma}_{s,t}, \bm{\beta}_{s,t}$ is the affine transformation learned by $h_{i,j}$ with the learnable parameters $\bm{\theta} = \begin{bmatrix}\bm{W^1}& \bm{W^2}& \bm{W^k}& \bm{W^b}\end{bmatrix}$ and relation function $h^{*}(\cdot, \cdot)$
\begin{equation}
    \begin{bmatrix}\bm{\gamma}_{s,t} & \bm{\beta}_{s,t}\end{bmatrix} = \sigma(\sigma(h^{*}(\bm{V}^{s}_{i}, \bm{V}^{t}_{i}) \bm{W^{1}}) \bm{W^{2}})\begin{bmatrix}\bm{W^{k}}& \bm{W^{b}}\end{bmatrix},
\end{equation}
where $\sigma(\cdot)$ is the rectified linear unit. 
The difference between different relation-mining operations lies in the choice of relation functions $h^{*}(\cdot, \cdot)$. 
We design 8 candidate relation functions, \ie, \emph{E\_SUB}, \emph{E\_GAUSS}, \emph{E\_HAD}, \emph{E\_MAX}, \emph{E\_SUM}, \emph{E\_MEAN}, \emph{skip-connect}, and \emph{zero} for capturing different types of relational information. 

\noindent
\textbf{\emph{E\_SUB}} captures the relative change between two nodes. Its relation function is computed as 
\begin{equation}
h^{*}(\bm{V}^{s}_{i}, \bm{V}^{t}_{i}) = \bm{V}^{s}_{i} - \bm{V}^{t}_{i}
\end{equation}

\noindent
\textbf{\emph{E\_GAUSS}} measures the distance between the central node and its neighboring nodes:
\begin{equation}
h^{*}(\bm{V}^{s}_{i}, \bm{V}^{t}_{i}) = exp(\frac{|\bm{V}^{s}_{i} - \bm{V}^{t}_{i}|^2}{2\sigma})
\end{equation}

\noindent
\textbf{\emph{E\_HAD}} emphasizes on learning the commonalities between the central node and its neighbors: 
\begin{equation}
h^{*}(\bm{V}^{s}_{i}, \bm{V}^{t}_{i}) = \bm{V}^{s}_{i} \odot \bm{V}^{t}_{i}
\end{equation}

\noindent
\textbf{\emph{E\_SUM}}:
\begin{equation}
h^{*}(\bm{V}^{s}_{i}, \bm{V}^{t}_{i}) = \bm{V}^{s}_{i} + \bm{V}^{t}_{i}
\end{equation}

\noindent
\textbf{\emph{E\_MAX}}:
\begin{equation}
h^{*}(\bm{V}^{s}_{i}, \bm{V}^{t}_{i}) = \max(\bm{V}^{s}_{i}, \bm{V}^{t}_{i})
\end{equation}

\noindent
\textbf{\emph{E\_MEAN}}:
\begin{equation}
h^{*}(\bm{V}^{s}_{i}, \bm{V}^{t}_{i}) = \frac{ \bm{V}^{s}_{i} + \bm{V}^{t}_{i} }{2}
\end{equation}

\noindent
\textbf{\emph{skip-connect}} operation is to enhance the original relational information and mitigate the gradient vanishing, written as 
\begin{equation}
    \begin{bmatrix}\bm{\gamma}_{s,t} & \bm{\beta}_{s,t}\end{bmatrix} = \begin{bmatrix}\bm{1} & \bm{0}\end{bmatrix}
\end{equation}

\noindent
\textbf{\emph{zero}} operation is included in the search space to indicate a lack of connection. 
Links that are important should have a low weight in the \emph{zero} operation \cite{SGAS}. 
It is formulated as
\begin{equation}
    \begin{bmatrix}\bm{\gamma}_{s,t} & \bm{\beta}_{s,t}\end{bmatrix} = \begin{bmatrix}\bm{0} & \bm{0}\end{bmatrix}
\end{equation}

\subsection{Task-based Layer}
We design the final network layers depending on the specific task. 
Suppose there are 8 node feature vertices $\{\bm{V}_{1}, \cdots, \bm{V}_{8}\}$ and 8 relation feature vertices $\{\bm{E}_{1}, \cdots, \bm{E}_{8}\}$ in our GNN architecture, we compute the global node features $\bm{V}_{g}$ and global edge features $\bm{E}_{g}$ using the following formula
\begin{gather}
    \begin{split}
        \bm{V}_{g} &= \sigma(BN(\left[\bm{V}_{1} || \cdots || \bm{V}_{8}\right]\bm{W}_{V})), \\
        \bm{E}_{g} &= \sigma(BN(\left[\bm{E}_{1} || \cdots || \bm{E}_{8}\right]\bm{W}_{E})), 
    \end{split}
\end{gather}
where $\bm{W}_{V} \in \mathbb{R}^{d_{V} \times d_{V}}, \bm{W}_{E} \in \mathbb{R}^{d_{E} \times d_{E}}$ are the learnable parameters, BN denotes batch normalization operation, $\sigma(\cdot)$ is the rectified linear unit, $[\cdot || \cdot]$ denotes the feature concatenation operation. 
The global graph representation $\bm{G}_{g}$ is computed using mean-pooling readout operation over global node features and global edge features, \ie, 
\begin{equation}
    \bm{G}_{g} = \left[\frac{1}{|\bm{V}_{g}|}\sum\limits_{i\in\bm{V}_{g}} \bm{V}_{g}^{i} \bigg| \bigg|  \frac{1}{|\bm{E}_{g}|}\sum\limits_{(s,t) \in\bm{E}_{g}}\bm{E}_{g}^{s,t}\right],
\end{equation}
where $|\bm{V}_{g}|$ is the number of nodes, $|\bm{E}_{g}|$ is the number of edges. 
Notably, this is different from the traditional GNN architecture whose global graph representation is only constructed on the readout of node features. 
Since our method can learn both node features and relation features, they are all leveraged to construct the global graph representation. 
This helps the graph representation embeds more useful relational information. 

\noindent
\textbf{Node-level task layer. }
For node classification task, the prediction of node $i$ is done as follows
\begin{equation}
\bm{y}^{i} = \bm{V}_{g}^{i}\bm{C}_{V},
\end{equation}
where $\bm{C}_{V} \in \mathbb{R}^{d_{V} \times n_V}$ is the node classifier, $n_V$ is the number of node classes.

\noindent
\textbf{Edge-level task layer. }
For edge classification task, our method naturally makes prediction based on deep relation features $\bm{E}_{g}$, formally written as 
\begin{equation}
    \bm{y}^{s,t} = \bm{E}_{g}^{s,t}\bm{C}_{E},
\end{equation}
where $\bm{C}_{E} \in \mathbb{R}^{d_{E} \times n_E}$ is the edge classifier, $n_E$ is the number of edge classes. 
This is better than traditional GNN works \cite{GCN, GAT, GIN, bench} that concatenate the entity features as edge features, since the independent edge feeatures (associated with the relational information) are more discriminitive for edge-level tasks.

\noindent
\textbf{Graph-level task layer. }
For graph classification and regression tasks, we make the prediction based on the global graph representation, \ie, 
\begin{equation}
    \bm{y} = \bm{G}_{g}\bm{C}_{G},
\end{equation}
where $\bm{C}_{G} \in \mathbb{R}^{(d_{V} + d_{E}) \times n_{G}}$, $n_{G}$ is the number of graph classses. 
If it is graph regression task, then $n_{G} = 1$.

\section{Network Differentiation Details}
An architecture is represented as a directed acyclic graph (DAG) with $\{\mathbb{V}, \mathbb{L}\}$, where $\mathbb{V} = \{\bm{X}_{i}\}$ is the set of feature vertices and $\mathbb{L} = \{e(\bm{X}_{i}, \bm{X}_{j}, O)\}$ is the set of directed links. 
Each directed link $e(\bm{X}_{i}, \bm{X}_{j}, O)$ can transform the features from $\bm{X}_{i}$ to $\bm{X}_{j}$ using an operation $O$, where $O$ is either a specific operation $\bm{o}$ or a mixture operation $\bm{\bar{o}}$. 
The mixture operation $\bm{\bar{o}}^{(i, j)}$ is parameterized by architectural parameters $\bm{\alpha}^{(i, j)}$ as a softmax mixture over all the possible operations within the operation space $\mathcal{O}$, \ie, 
\begin{equation}
\bm{\bar{o}}^{(i, j)}(X_{i}) = \sum_{\bm{o} \in \mathcal{O}} \frac{exp(\alpha^{(i, j)}_{o})}{\sum_{o' \in \mathcal{O}} exp(\alpha^{(i, j)}_{o'})} \bm{o}(X_{i}). 
\end{equation}
For each operation $\bm{o}^{(i,j)}$, it is associated with the network weights $\bm{w}^{(i,j)}$. 

The network differentiation aims to differentiate the local supernets into several specific subnets. 
Specifically, after network division and before network differentiation, the current architecture contains two kinds of links $e(\bm{X}_{i}, \bm{X}_{j}, \bm{o})$ and $e(\bm{X}_{i}, \bm{X}_{j}, \bm{\bar{o}})$. 
After network differentiation, there is only one kind of links, \ie, $e(\bm{X}_{i}, \bm{X}_{j}, \bm{o})$, where a valid architecture is obtained. 
This procedure can be implemented by some differentiable architecture search strategies (DARTS \cite{DARTS}, SGAS \cite{SGAS}, \etc). 
Taking DARTS as an example, the learning procedure of the architectural parameters involves a bi-level optimization problem, \ie, 
\begin{gather}
\min\limits_{\mathcal{A}} \mathcal{L}_{val}(\mathcal{W}^{*}(\mathcal{A}), \mathcal{A}), \\
s.t. \mathcal{W}^{*}(\mathcal{A}) = \mathop{argmin}\limits_{\mathcal{W}} \mathcal{L}_{train}(\mathcal{W}, \mathcal{A}),
\end{gather}
where $\mathcal{L}_{train}$ and $\mathcal{L}_{val}$ are the training and validation loss, respectively. 
$\mathcal{W}$ is the set of network weigths $\{\bm{w}^{(i, j)}\}$ and $\mathcal{A}$ is the set of the architectural parameters $\{\bm{\alpha}^{(i, j)}\}$. 
DARTS \cite{DARTS} proposes to solve the bi-level problem by a first/second order approximation. 
At the end of the search, the final architecture is derived by selecting the operation with the highest weight for each mixture operation, \ie, 
\begin{equation}
    \bm{\bar{o}}^{(i, j)} \leftarrow \mathop{argmax}_{\bm{o} \in \mathcal{O}} \alpha^{(i, j)}_{\bm{o}}. 
\end{equation}

\section{Visualizing Hierarchical Features}
\begin{figure*}[t]
    \centering 
    \includegraphics[scale = 1.05, trim = 80 140 80 140, clip]{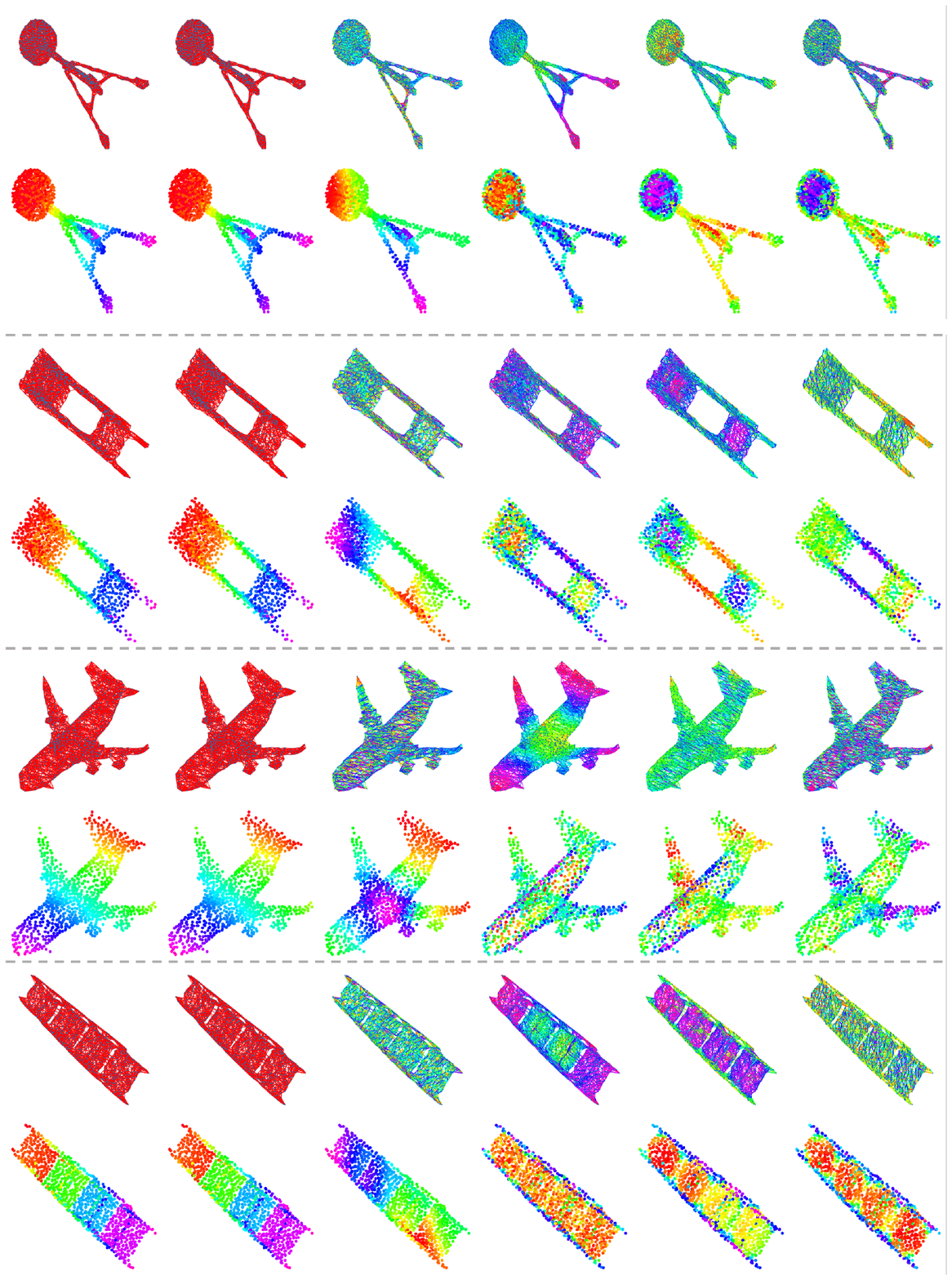}
    \caption{
        Visualization of the learned hierarchical relation and node features for 3D point cloud recognition. 
        Relation features with different edge color distributions have different message-passing preferences. 
        Node features with different node color distributions represent different clustering effects. 
    }
    \label{pc_supp}
\end{figure*}

In this section, we provide more examples of the learned relation and node features on ModelNet40. 
The visualization results are reported in Figure \ref{pc_supp}.

\section{Searched Architectures}

\begin{figure*}[t]
    \centering 
    \includegraphics[width=0.6\textwidth]{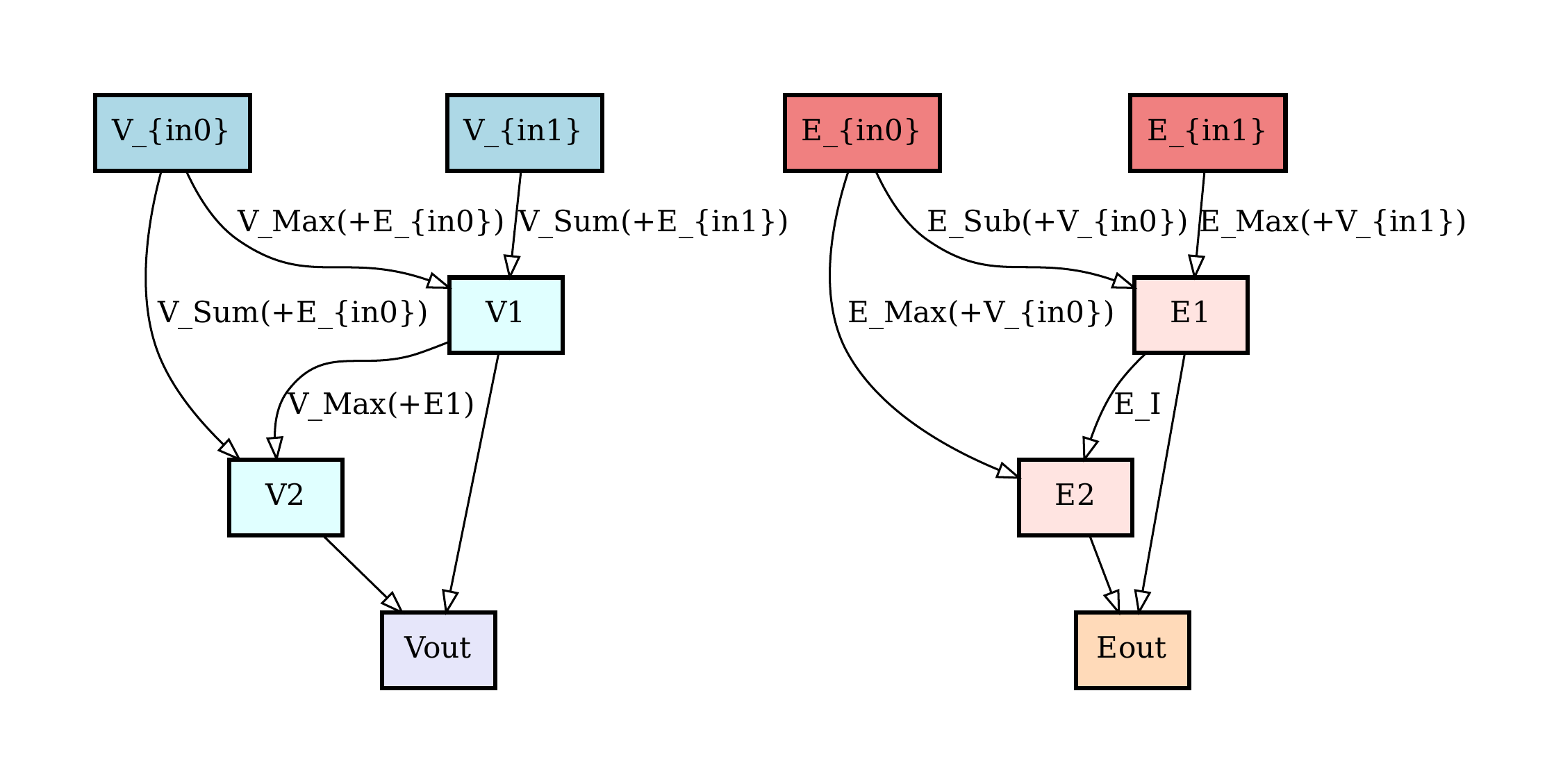}
    \caption{Illustration of the searched architecture with the size of 2 on the ZINC dataset. }
    \label{f1}
\end{figure*}

\begin{figure*}[t]
    \centering 
    \includegraphics[width=1.0\textwidth]{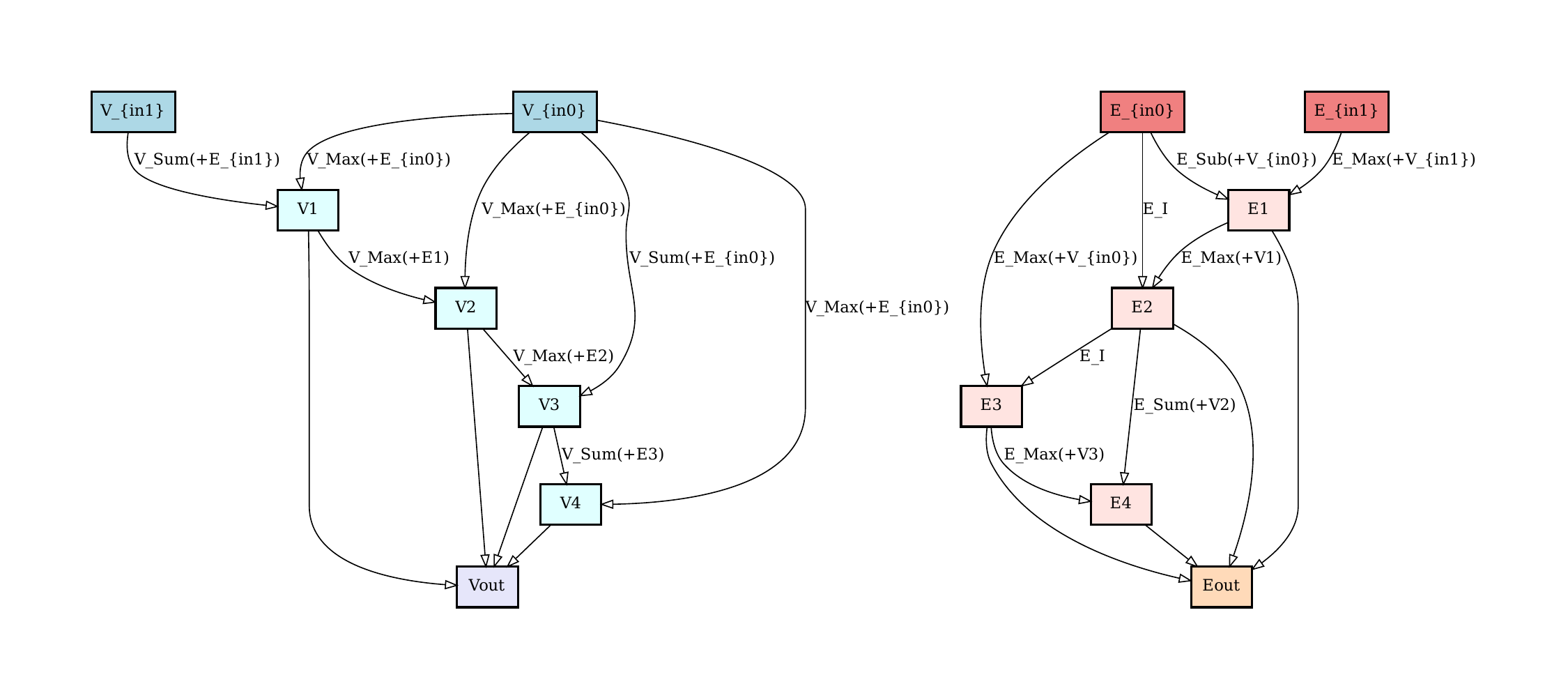}
    \caption{Illustration of the searched architecture with the size of 4 on the ZINC dataset. }
    \label{f2}
\end{figure*}

\begin{figure*}[t]
    \centering 
    \includegraphics[width=1.0\textwidth]{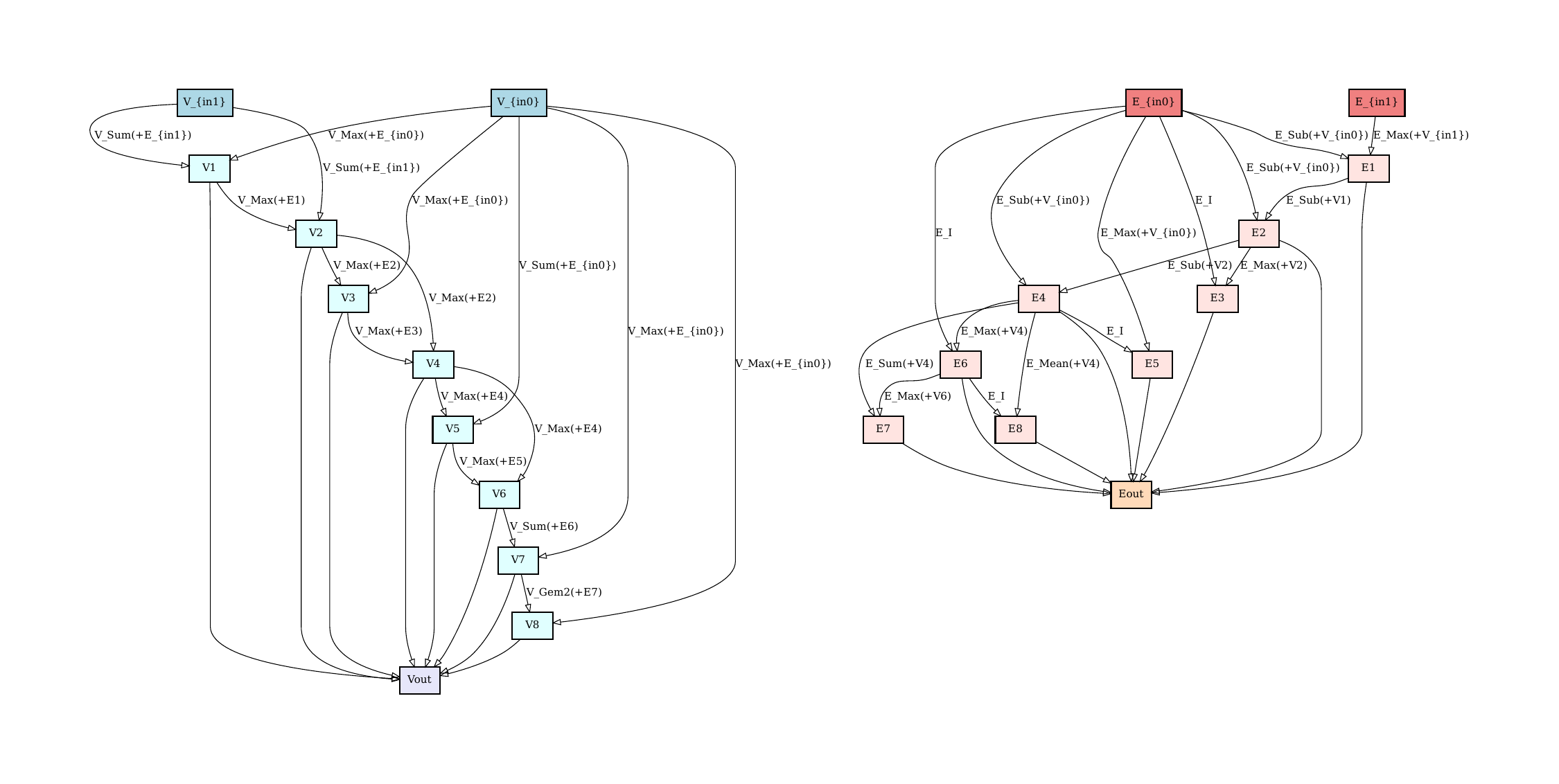}
    \caption{Illustration of the searched architecture with the size of 8 on the ZINC dataset. }
    \label{f3}
\end{figure*}

\begin{figure*}[t]
    \centering 
    \includegraphics[width=1.0\textwidth]{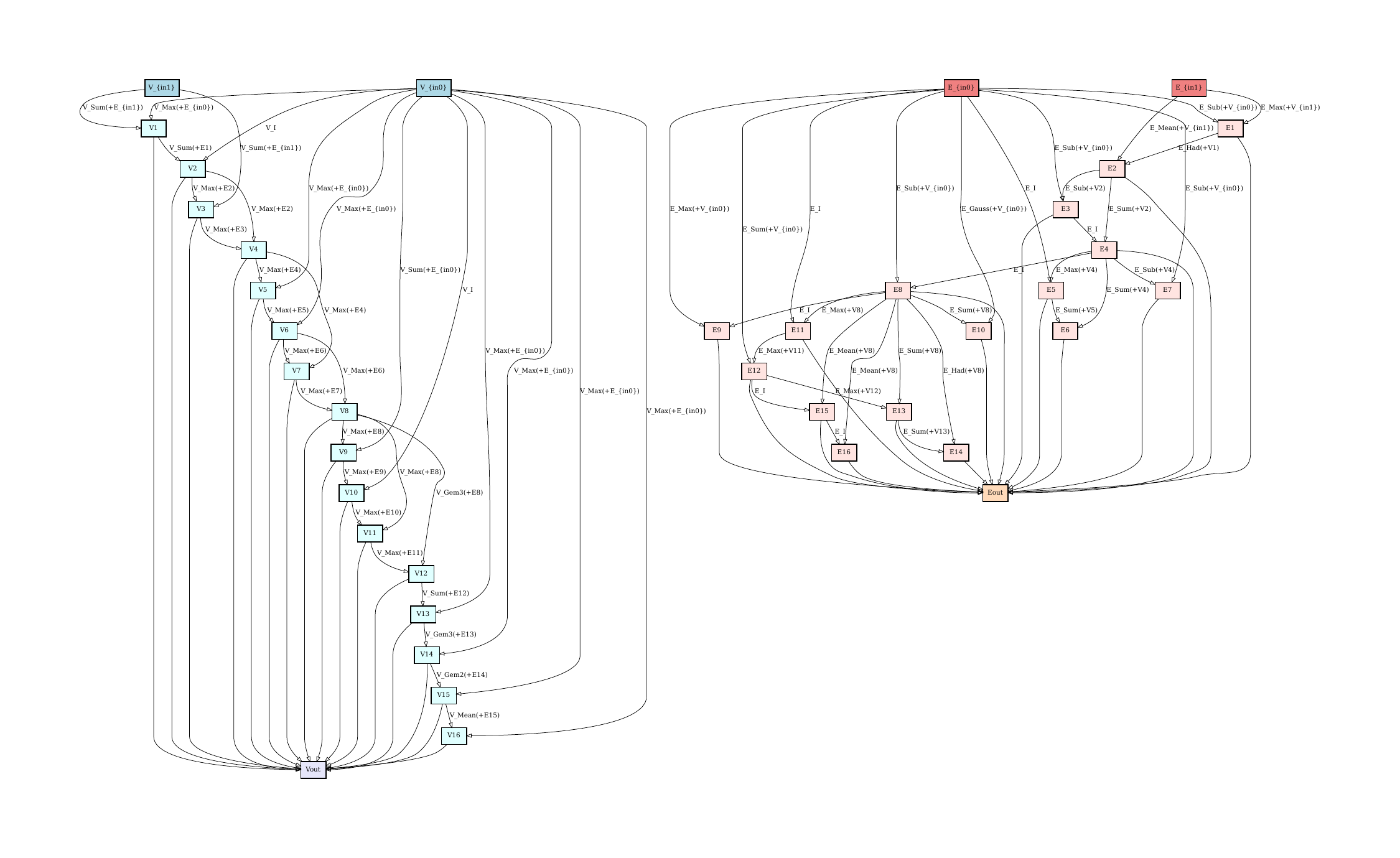}
    \caption{Illustration of the searched architecture with the size of 16 on the ZINC dataset. }
    \label{f4}
\end{figure*}

\begin{figure*}[t]
    \centering 
    \includegraphics[width=0.8\textwidth]{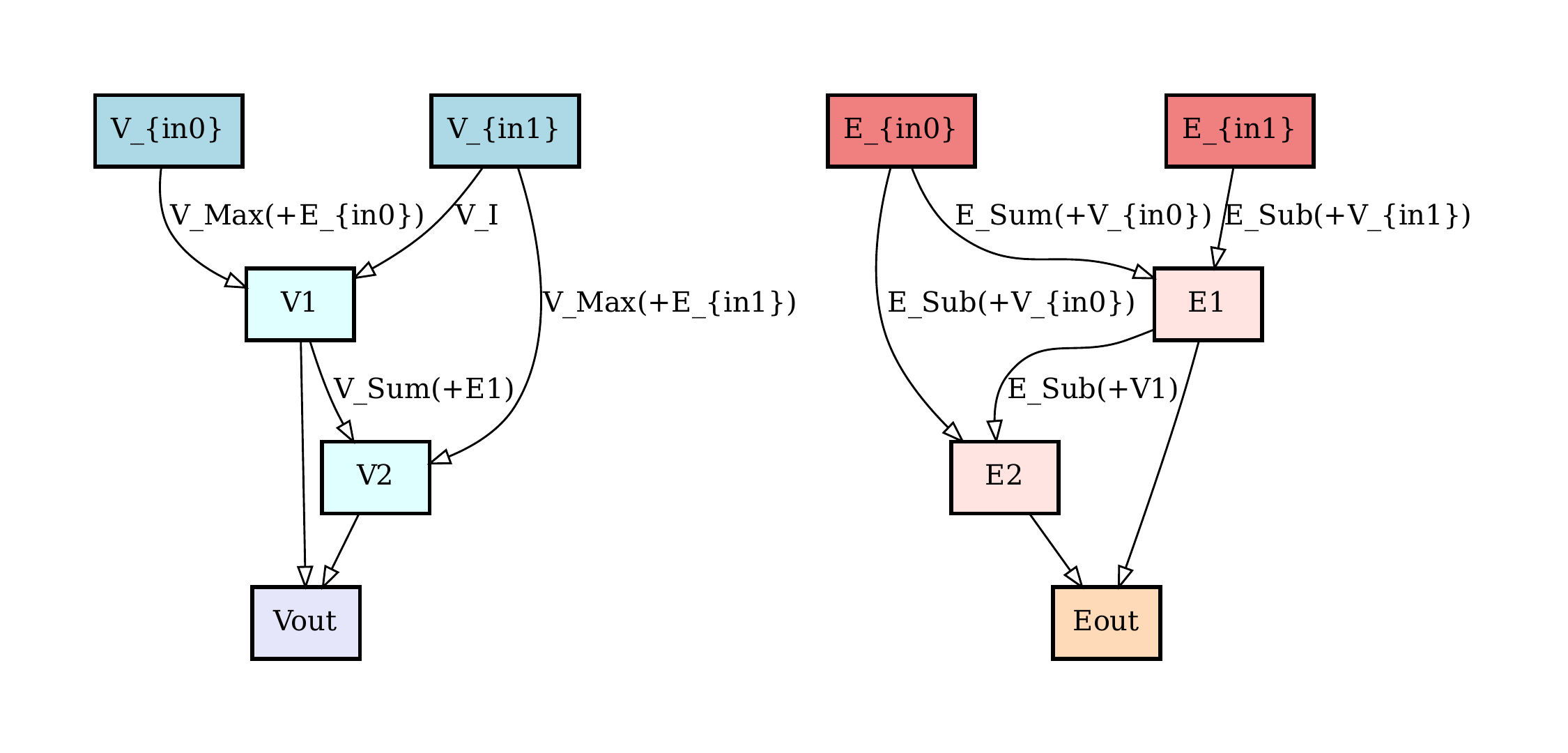}
    \caption{Illustration of the searched architecture with the size of 2 on the CLUSTER dataset. }
    \label{f5}
\end{figure*}

\begin{figure*}[t]
    \centering 
    \includegraphics[width=1.0\textwidth]{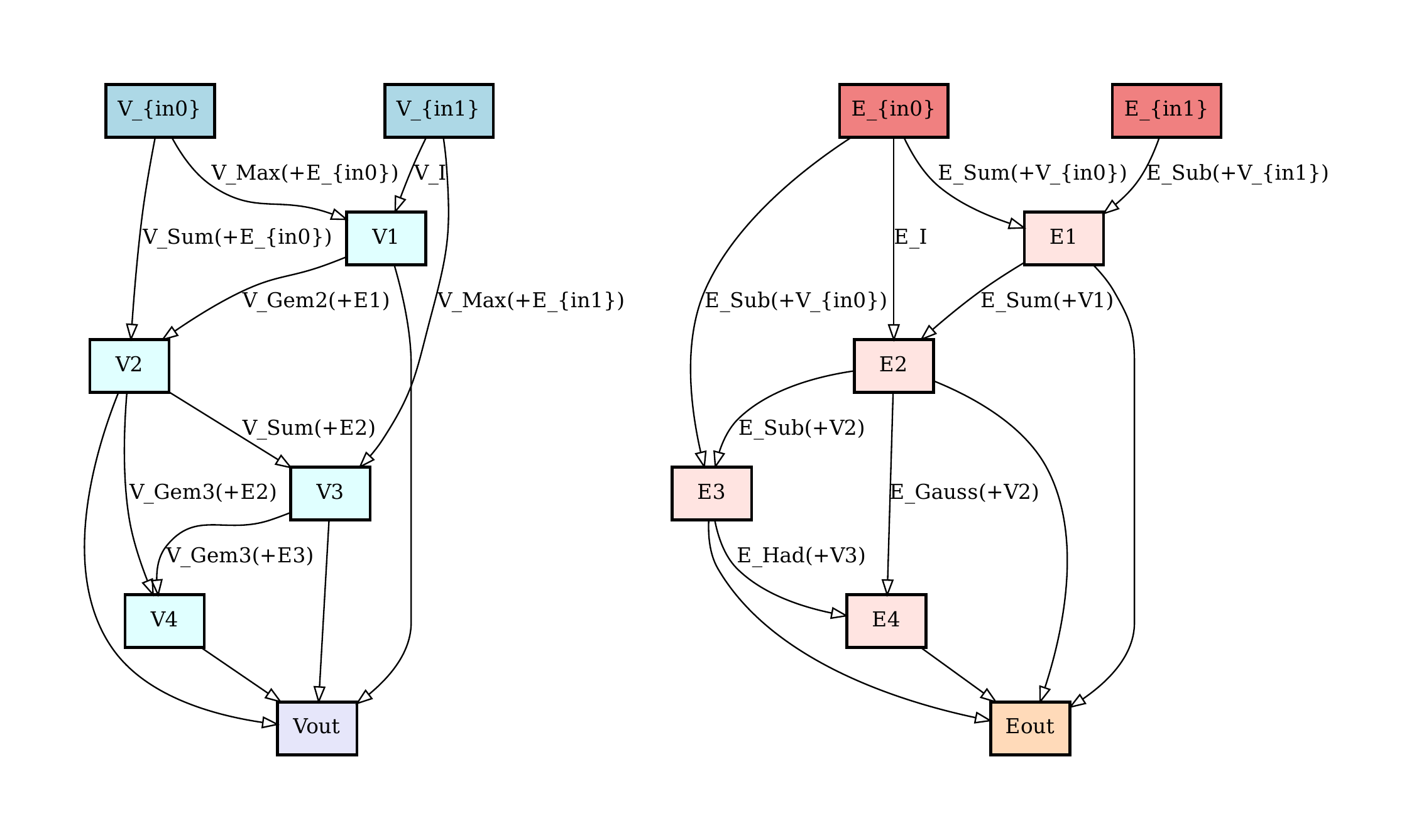}
    \caption{Illustration of the searched architecture with the size of 4 on the CLUSTER dataset. }
    \label{f6}
\end{figure*}

\begin{figure*}[t]
    \centering 
    \includegraphics[width=0.8\textwidth]{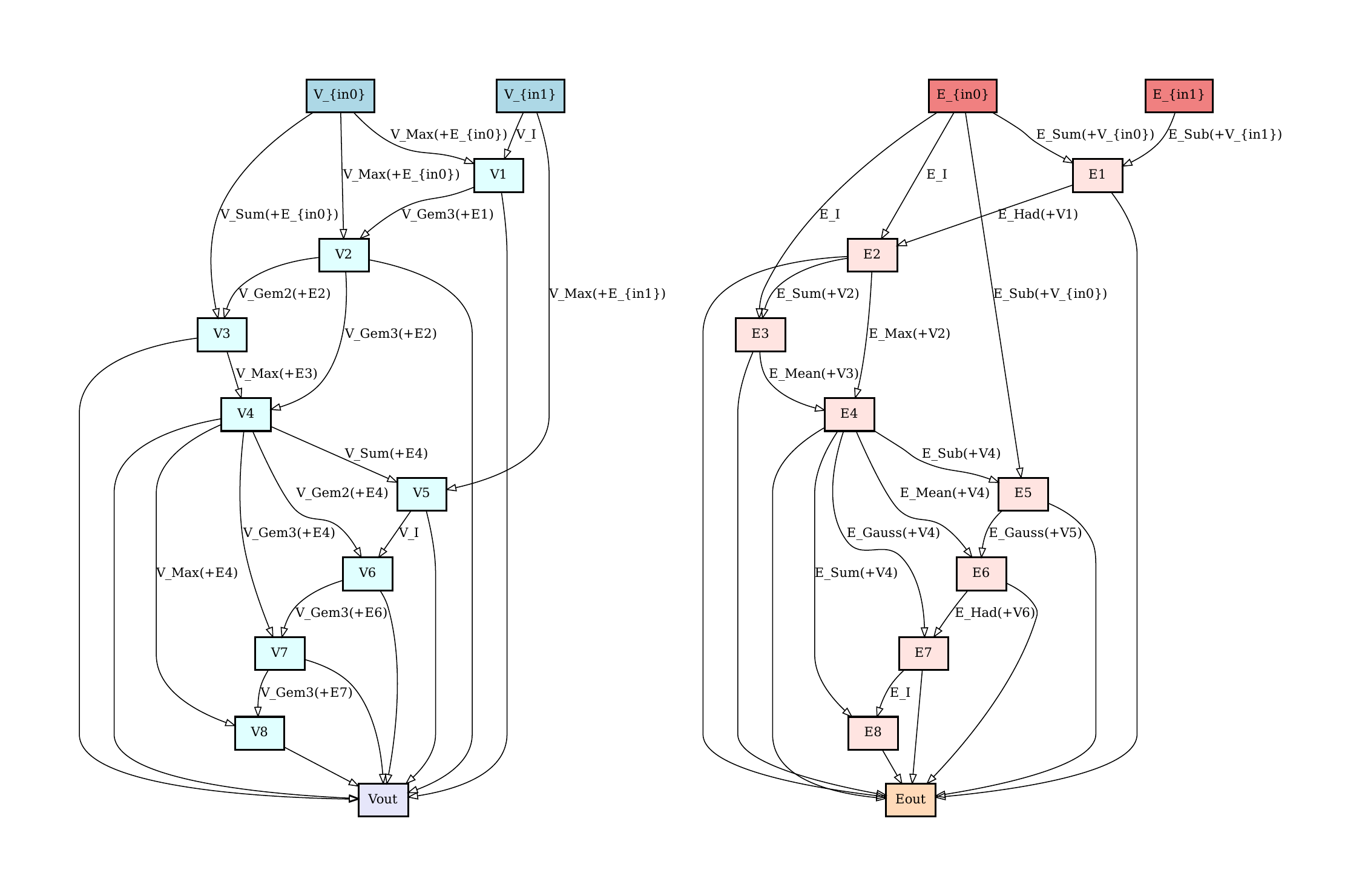}
    \caption{Illustration of the searched architecture with the size of 8 on the CLUSTER dataset. }
    \label{f7}
\end{figure*}

\begin{figure*}[t]
    \centering 
    \includegraphics[width=1.0\textwidth]{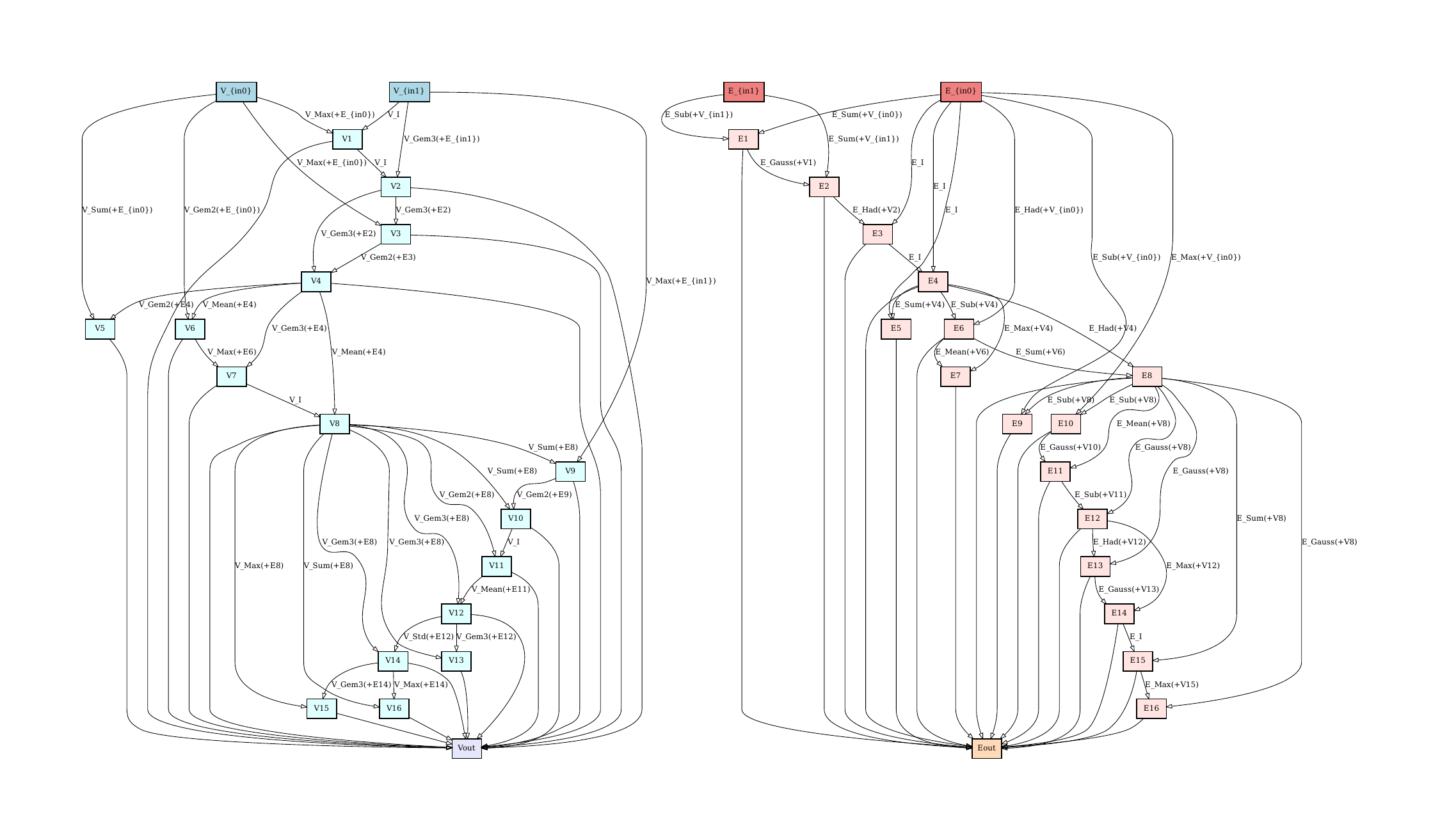}
    \caption{Illustration of the searched architecture with the size of 16 on the CLUSTER dataset. }
    \label{f8}
\end{figure*}

\begin{figure*}[t]
    \centering 
    \includegraphics[width=0.6\textwidth]{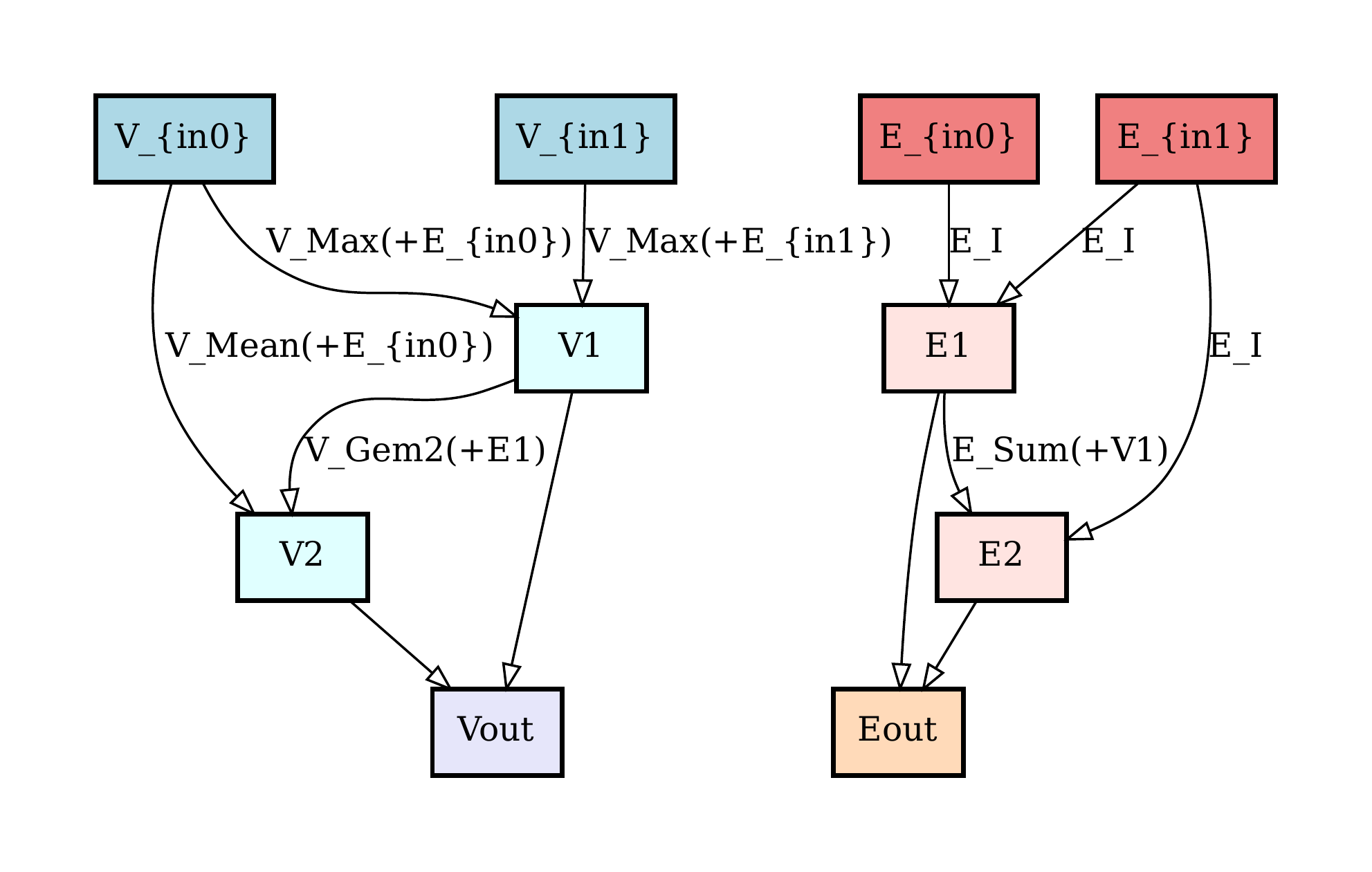}
    \caption{Illustration of the searched architecture with the size of 2 on the TSP dataset. }
    \label{f9}
\end{figure*}

\begin{figure*}[t]
    \centering 
    \includegraphics[width=1.0\textwidth]{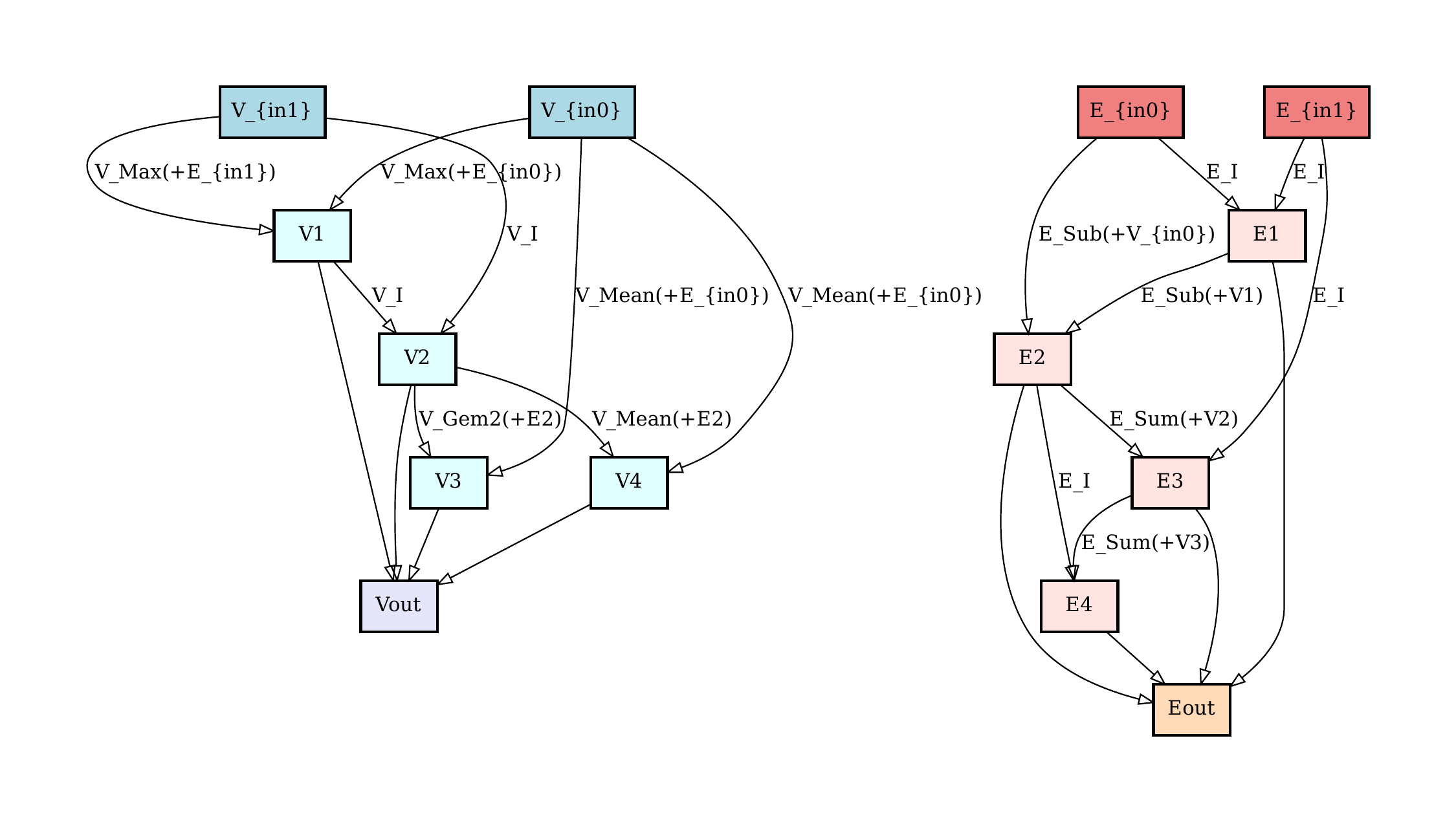}
    \caption{Illustration of the searched architecture with the size of 4 on the TSP dataset. }
    \label{f10}
\end{figure*}

\begin{figure*}[t]
    \centering 
    \includegraphics[width=0.8\textwidth]{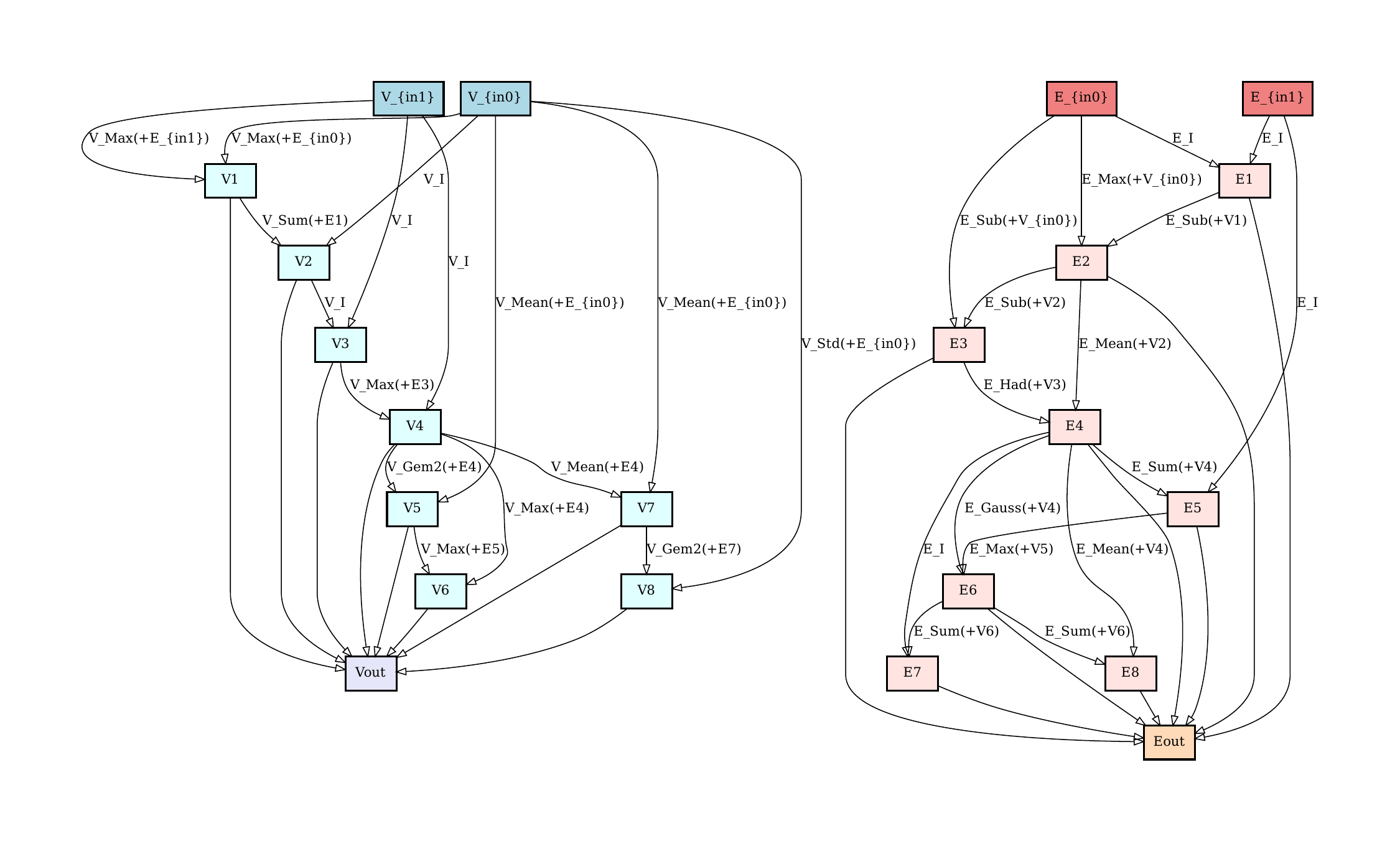}
    \caption{Illustration of the searched architecture with the size of 8 on the TSP dataset. }
    \label{f11}
\end{figure*}

\begin{figure*}[t]
    \centering 
    \includegraphics[width=1.0\textwidth]{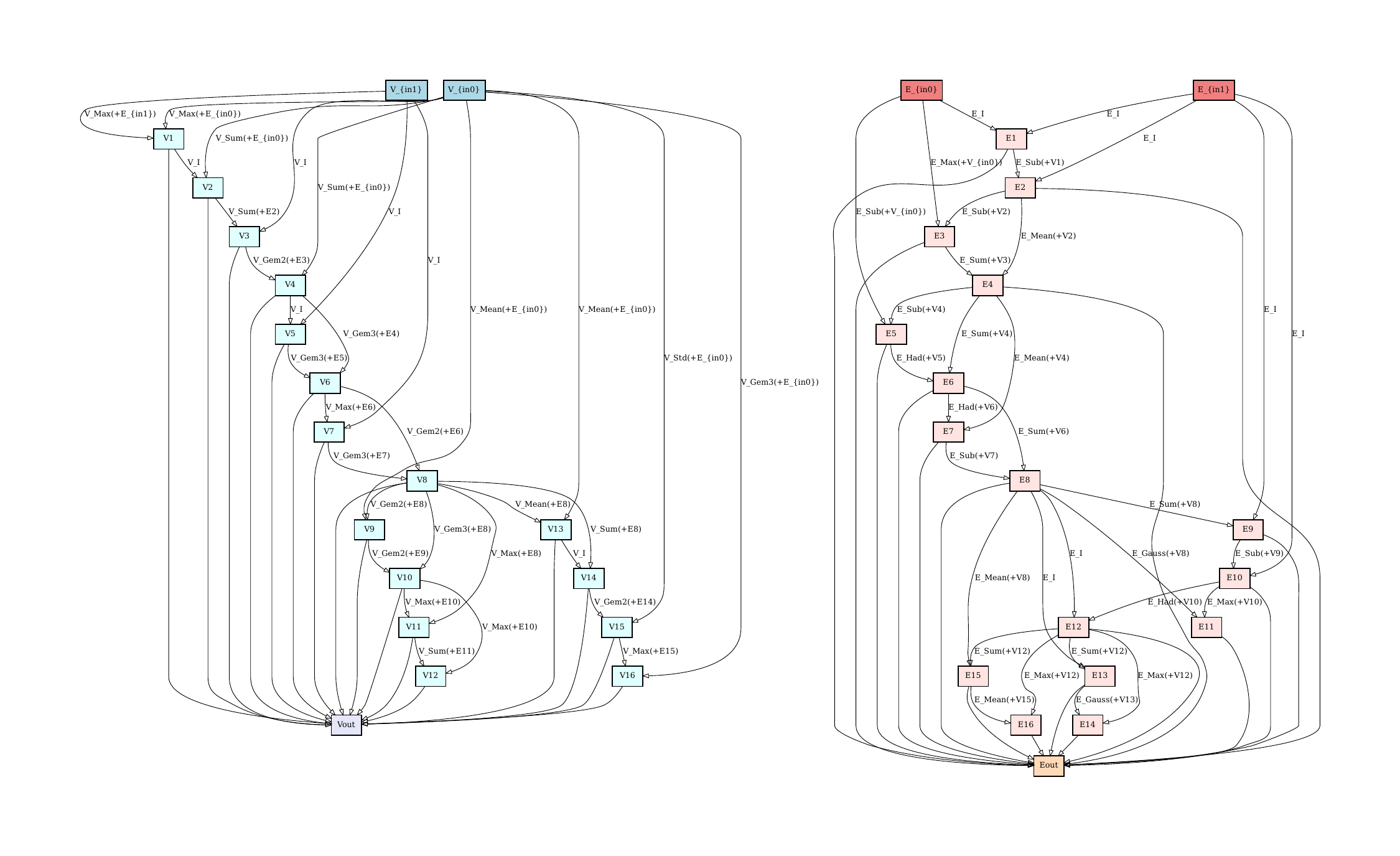}
    \caption{Illustration of the searched architecture with the size of 16 on the TSP dataset. }
    \label{f12}
\end{figure*}

\begin{figure*}[t]
    \centering 
    \includegraphics[width=1.0\textwidth]{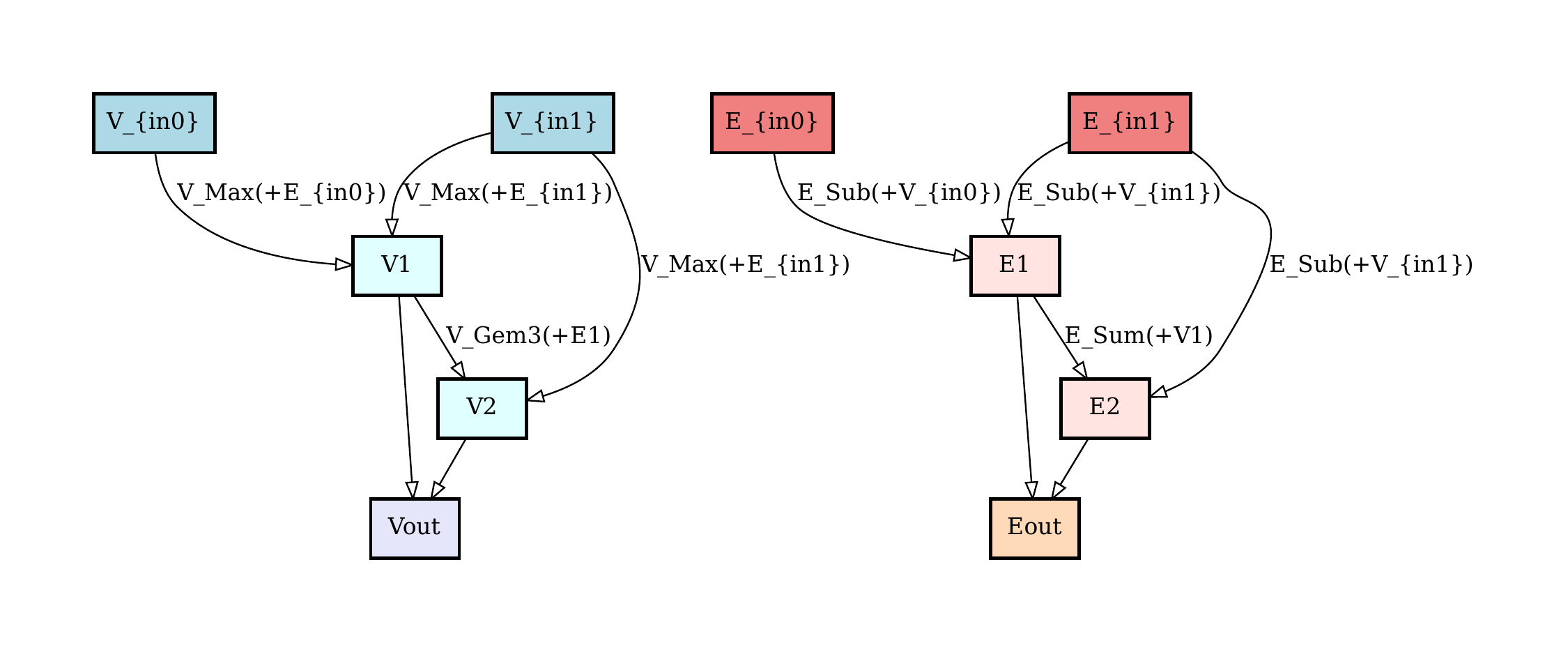}
    \caption{Illustration of the searched architecture with the size of 2 on the CIFAR10 dataset. }
    \label{f13}
\end{figure*}

\begin{figure*}[t]
    \centering 
    \includegraphics[width=1.0\textwidth]{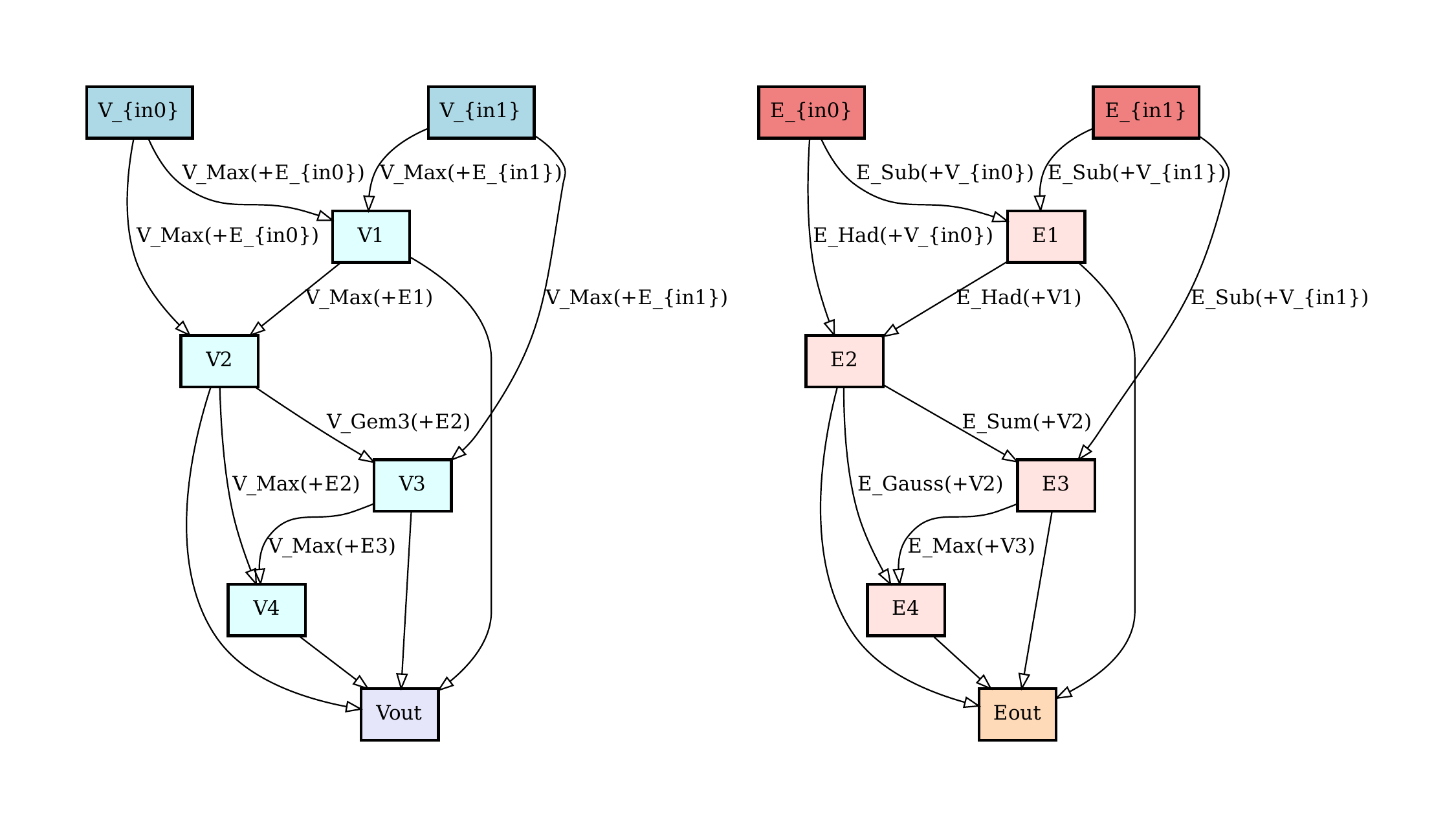}
    \caption{Illustration of the searched architecture with the size of 4 on the CIFAR10 dataset. }
    \label{f14}
\end{figure*}

\begin{figure*}[t]
    \centering 
    \includegraphics[width=1.0\textwidth]{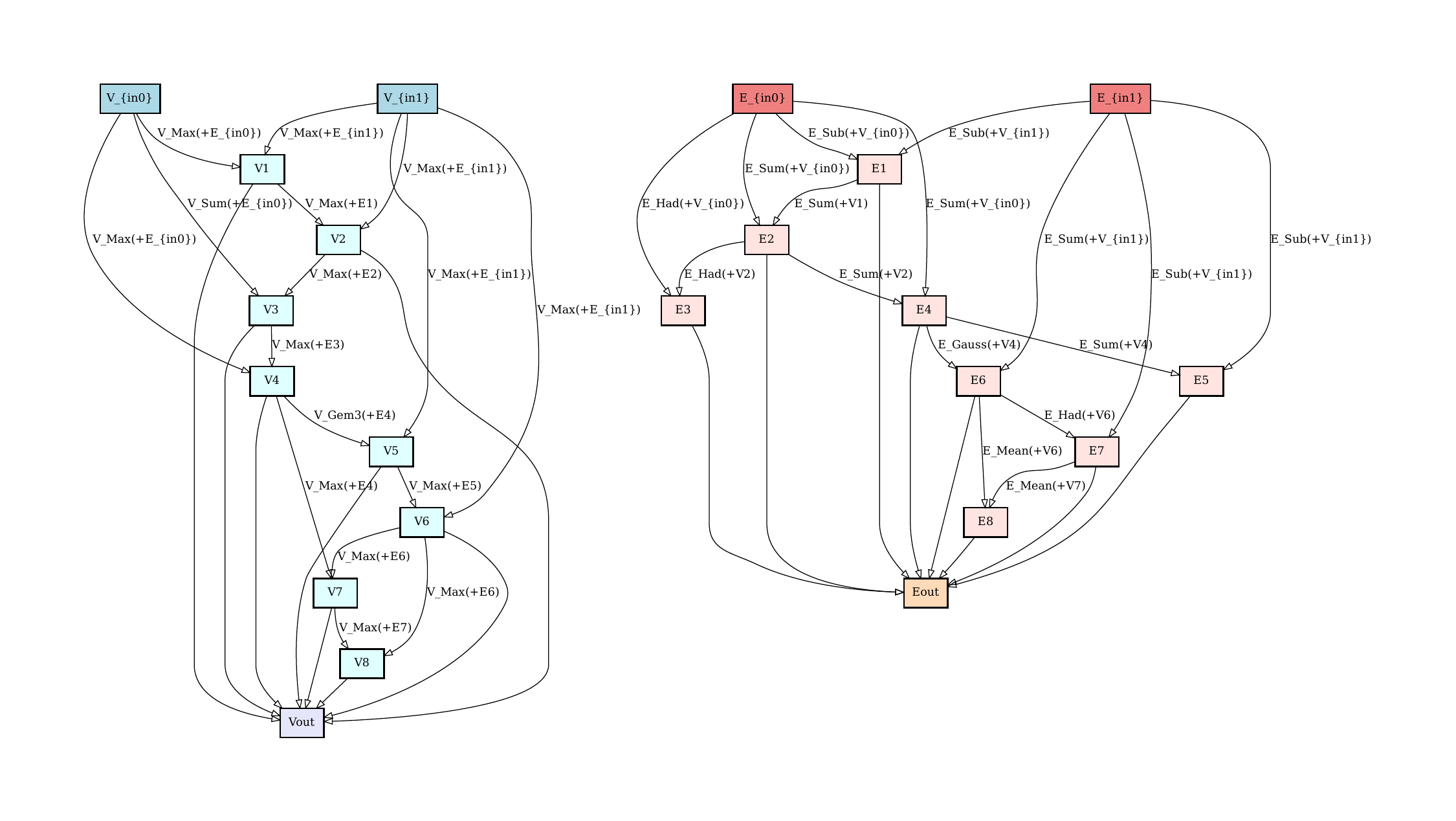}
    \caption{Illustration of the searched architecture with the size of 8 on the CIFAR10 dataset. }
    \label{f15}
\end{figure*}

\begin{figure*}[t]
    \centering 
    \includegraphics[width=1.1\textwidth]{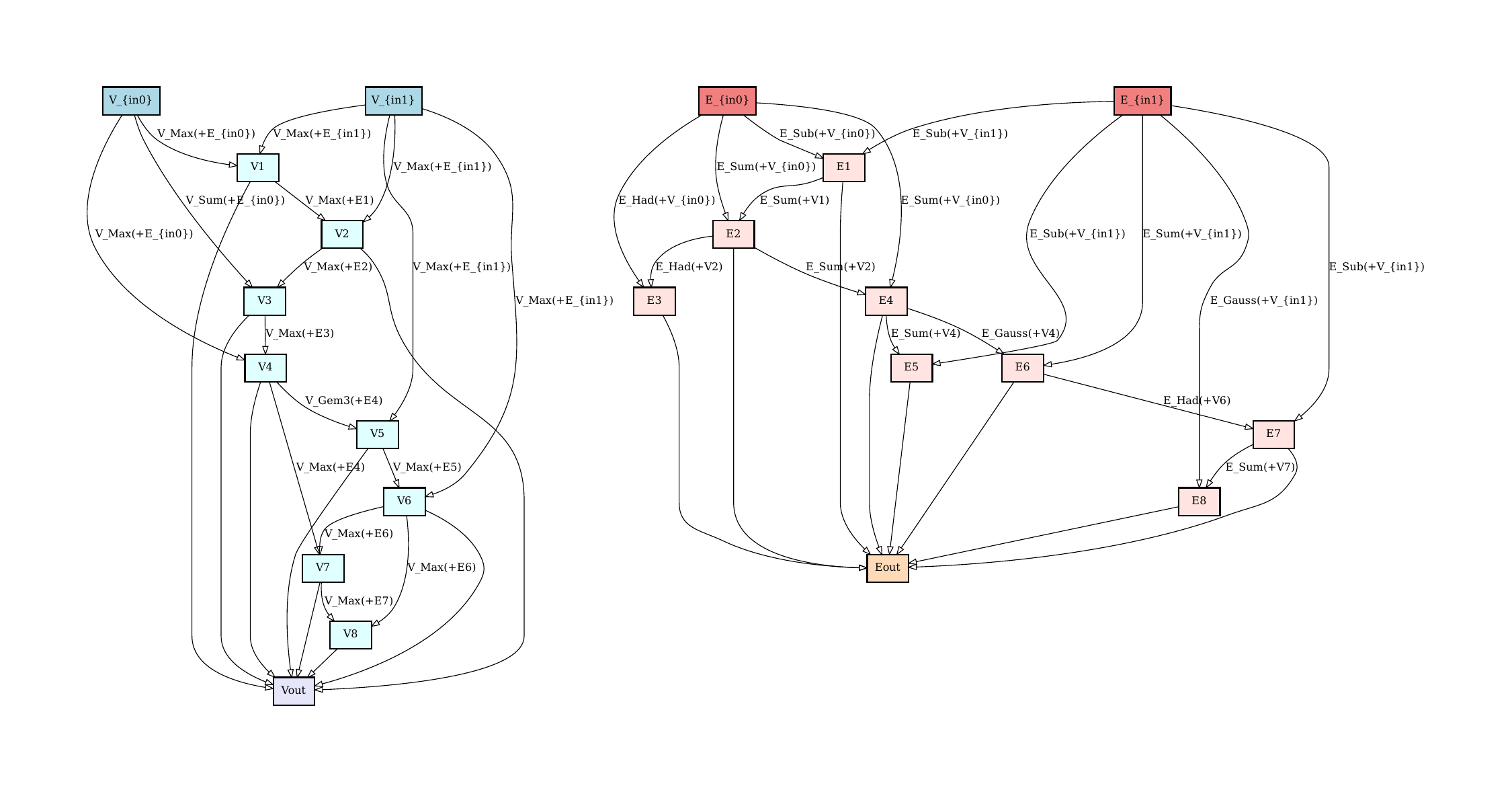}
    \caption{Illustration of the searched architecture with the size of 16 on the CIFAR10 dataset. }
    \label{f16}
\end{figure*}

\begin{figure*}[t]
    \centering 
    \includegraphics[width=1.0\textwidth]{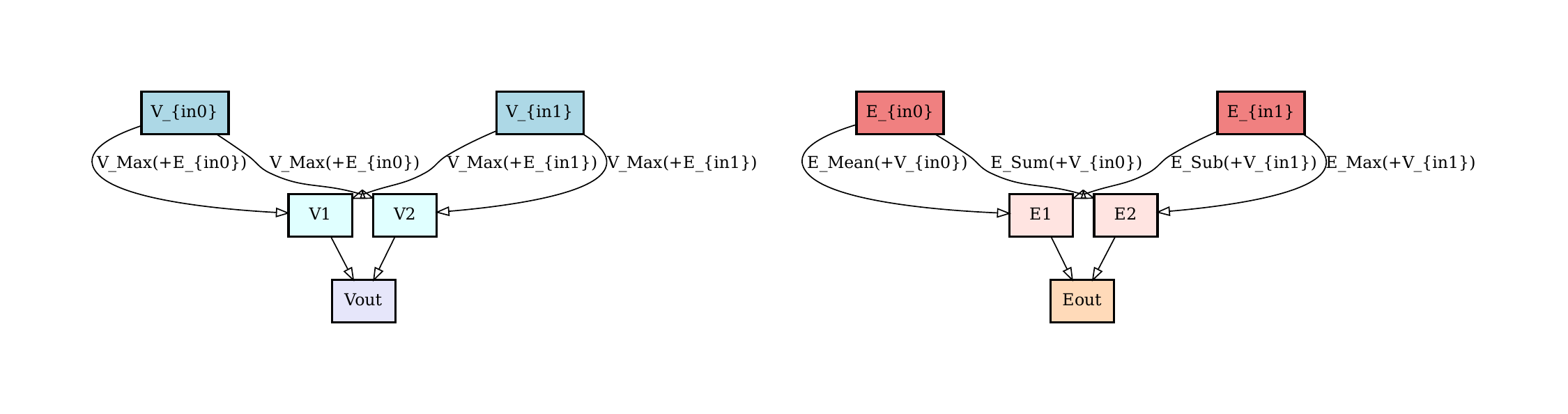}
    \caption{Illustration of the searched architecture with the size of 2 on the ModelNet10 dataset. }
    \label{f17}
\end{figure*}

\begin{figure*}[t]
    \centering 
    \includegraphics[width=1.0\textwidth]{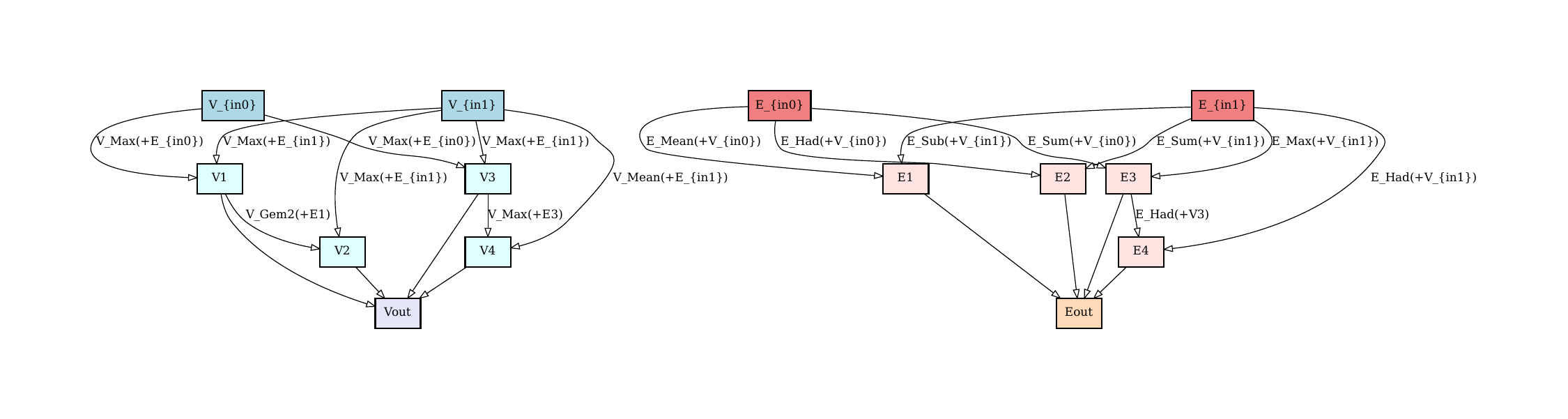}
    \caption{Illustration of the searched architecture with the size of 4 on the ModelNet10 dataset. }
    \label{f18}
\end{figure*}

\begin{figure*}[t]
    \centering 
    \includegraphics[width=1.0\textwidth]{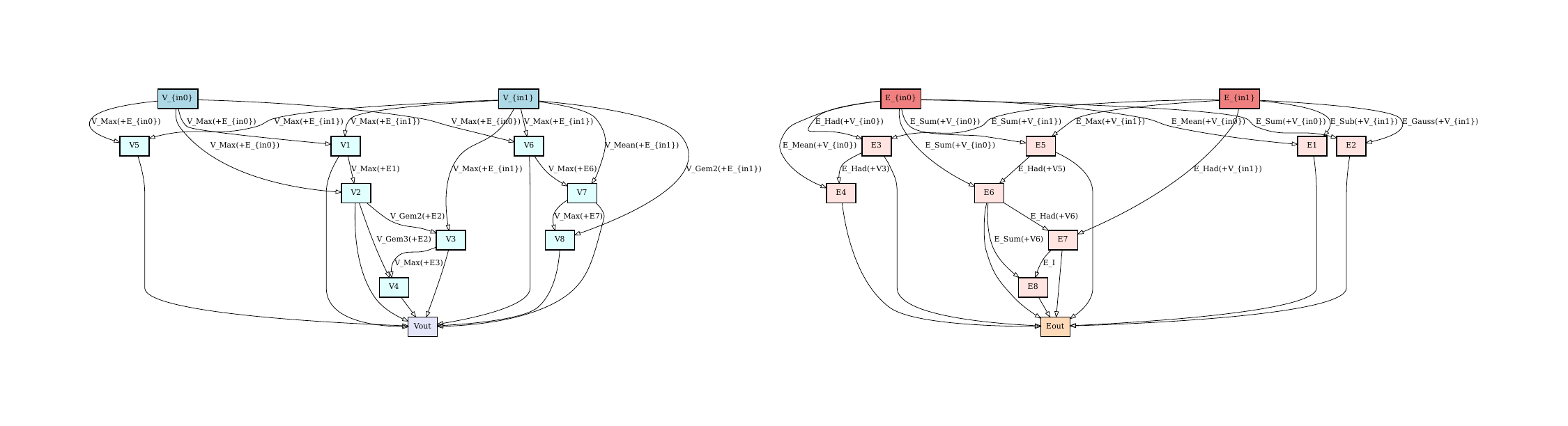}
    \caption{Illustration of the searched architecture with the size of 8 on the ModelNet10 dataset. }
    \label{f19}
\end{figure*}

We illustrate our searched GNN architectures using the proposed \emph{network proliferation search paradigm} in Figure \ref{f1}, \ref{f2}, \ref{f3}, \ref{f4}, \ref{f5}, \ref{f6}, \ref{f7}, \ref{f8}, \ref{f9}, \ref{f10}, \ref{f11}, \ref{f12}, \ref{f13}, \ref{f14}, \ref{f15}, \ref{f16}, \ref{f17}, \ref{f18}, \ref{f19}.

\end{document}